\newif\ifsingle
\singletrue 

\newif\ifFullVersion
\FullVersiontrue 

\ifsingle
\documentclass[11pt,draftclsnofoot, onecolumn]{IEEEtran}		
\else		
\documentclass[10pt,final, twocolumn]{IEEEtran}
\fi

\usepackage{times}
\usepackage{amsmath,dsfont}
\usepackage{amssymb,amsthm}
\usepackage{epsfig,verbatim} 
\usepackage{setspace}
\usepackage{color}
\usepackage{cite}
\usepackage{epstopdf}
\usepackage{graphics}
\usepackage{accents}
\usepackage{acronym}
\usepackage[bookmarks,colorlinks]{hyperref}
\usepackage{booktabs}
\usepackage{mathtools}
\usepackage{algorithm}
\usepackage{algorithmic}
\usepackage{enumitem}
\usepackage{wrapfig}
\usepackage{subcaption}

\ifsingle
\newcommand{\figWidth}{0.65\columnwidth}
\newcommand{\figHeight}{2.9in}
\newcommand{\figSpace}{\vspace{-0.2cm}}
\newcommand{\includefig}[1]{\includegraphics[width = 0.75\columnwidth]{#1} 	\vspace{-0.2cm}}
\else
\newcommand{\figWidth}{\columnwidth}
\newcommand{\figHeight}{2.7in}
\newcommand{\includefig}[1]{\includegraphics[width = \columnwidth]{#1} 	\vspace{-0.2cm}}
\newcommand{\figSpace}{\vspace{-0.4cm}}
\fi 

\definecolor{NewColor}{rgb}{0,0,0}

\newcommand{\myVec}[1]{{\boldsymbol{#1}}}
\newcommand{\myMat}[1]{{\boldsymbol{#1}}}
\newcommand{\mySet}[1]{\mathcal{#1}}
\newcommand{\E}{\mathds{E}}		 			
\newcommand{\myI}{{\myMat{I}}}			 		
\newcommand{\myX}{{\myVec{x}}}			 		
\newcommand{\myY}{\myVec{y}}
\newcommand{\myWeights}{\myVec{w}}
\newcommand{\myWeightsV}{\bar{\myWeights}}
\newcommand{\myUpdate}{\myVec{h}}
\newcommand{\myUpdateV}{\bar{\myVec{h}}}
\newcommand{\myGlobalModel}{\tilde{\myWeights}} 
\newcommand{\myGlobalModelV}{\bar{\myWeights}} 
\newcommand{\Nusers}{K}
\newcommand{\Ntraining}{n}
\newcommand{\Rate}{R}
\newcommand{\Nlattice}{L}
\newcommand{\Npoints}{M}
\newcommand{\GenMat}{\myMat{G}}
\newcommand{\Partition}{\mySet{P}_0}
\newcommand{\Desired}{^{\rm des}}
\newcommand{\Opt}{^{\rm o}}

\newcommand{\IdxK}{^{(k)}}
\newcommand{\IdxKt}{^{(k')}}
\newcommand{\SGDIter}{\tau}
\newcommand{\Objective}{F}
\newcommand{\SmoothParam}{{\rho}_s}
\newcommand{\ConvParam}{{\rho}_c}
\newcommand{\StepSize}{\eta}

\newcommand{\Qerror}{\myVec{\epsilon}}
\newcommand{\QerrorV}{\bar{\Qerror}}
\newcommand{\QerrorSeq}{\myVec{e}}

\newcommand{\GradDet}{\myVec{g}}
\newcommand{\GradRand}{\tilde{\GradDet}}
\newcommand{\GradMom}{\xi^2}
\newcommand{\GradVar}{\GradMom}
\newcommand{\HetMismatch}{\psi}

\newcommand{\LatFactor}{\bar{\sigma}_{\mySet{L}}^2}
\newcommand{\ScaleFactor}{\zeta}

\newtheorem{theorem}{Theorem}

\newtheorem{lemma}{Lemma}

\acrodef{adc}[ADC]{analog-to-digital convertor}
\acrodef{cs}[CS]{compressed sensing}
\acrodef{dtft}[DTFT]{discrete-time Fourier transform}
\acrodef{dnn}[DNN]{deep neural network} 
\acrodef{csi}[CSI]{channel state information}
\acrodef{map}[MAP]{maximum a-posteriori probability}
\acrodef{snr}[SNR]{signal-to-noise ratio}
\acrodef{bs}[BS]{base station} 
\acrodef{iot}[IOT]{Interent of Things}
\acrodef{mimo}[MIMO]{multiple-input multiple-output}
\acrodef{mse}[MSE]{mean-squared error}
\acrodef{pdf}[PDF]{probability density function}
\acrodef{rv}[RV]{random variable}
\acrodef{fec}[FEC]{forward error correction}
\acrodef{fl}[FL]{Federated learning}
\acrodef{lti}[LTI]{linear time-invariant}
\acrodef{wss}[WSS]{wide-sense stationary}
\acrodef{psd}[PSD]{power spectral density}
\acrodef{ser}[SER]{symbol error rate} 
\acrodef{ber}[BER]{bit error rate} 
\acrodef{sgd}[SGD]{stochastic gradient descent} 
\acrodef{isi}[ISI]{intersymbol interference}  
\acrodef{awgn}[AWGN]{additive white Gaussian noise} 
\acrodef{ut}[UT]{user terminal} 
\acrodef{uveqfed}[UVeQFed]{unviersal vector quantization for federated learning}

\IEEEoverridecommandlockouts

\title{UVeQFed: Universal Vector Quantization for Federated Learning
}
\author{
	\IEEEauthorblockN{Nir Shlezinger, Mingzhe Chen, Yonina C. Eldar, H. Vincent Poor, and Shuguang Cui\\
	} 
	\thanks{
		    Parts of this work were presented in the IEEE International Conference on Acoustics, Speech, and Signal Processing (ICASSP) 2020 as the paper \cite{shlezinger2020federated}. 
	The work of Y. C. Eldar was supported in part by the Benoziyo Endowment Fund for the Advancement of Science, the	Estate of Olga Klein -- Astrachan, the European Union’s Horizon 2020 research and innovation program under grant No. 646804-ERC-COG-BNYQ, and from the Israel Science Foundation under grant No. 0100101.
The work of H. V. Poor was supported in part by the U.S. National Science Foundation under grants CCF-0939370 and CCF-1908308. 
The work of S. Cui was supported in part by the Key Area R\&D Program of Guangdong Province with grant No. 2018B030338001, by the National Key R\&D Program of China with grant No. 2018YFB1800800, by Natural Science Foundation of China with grant NSFC-61629101, and by Guangdong Zhujiang Project No. 2017ZT07X152.
			}
			\thanks{N. Shlezinger is with the School of ECE, Ben-Gurion University of the Negev, Be'er-Sheva, Israel (e-mail:nirshl@bgu.ac.il).
		M. Chen  and H. V. Poor are with the EE Dept., Princeton University, Princeton, NJ  (e-mail: \{mingzhec, poor\}@princeton.edu). 	M. Chen is also with the Chinese University of Hong Kong, Shenzhen, China.
		Y. C. Eldar is with the Faculty of Math and CS, Weizmann Institute of Science, Rehovot, Israel (e-mail: yonina@weizmann.ac.il). 	
		S. Cui is with the Chinese University of Hong Kong, Shenzhen, China (e-mail: shuguangcui@cuhk.edu.cn)}

	\vspace{-1.0cm}
	
}
\vspace{-0.75cm}

\begin{document}
	
	\maketitle
	\pagestyle{plain}
	\thispagestyle{plain}
	
	\begin{abstract}
		Traditional deep learning models are trained at a centralized server using labeled data samples collected from end devices or users. Such data samples often include private information, which the  users may not be willing to share. Federated learning (FL) is an emerging approach to train such learning models without requiring the users to share their possibly private labeled data. FL consists of an iterative procedure, where in each iteration the users train a copy of the learning model locally. The server then collects the individual updates and aggregates them into a global model. A major challenge that arises in this method is the need of each user to repeatedly transmit its learned model over the throughput limited uplink channel. In this work, we tackle this challenge using tools from quantization theory. In particular, we identify the unique characteristics associated with conveying trained models over rate-constrained channels, and propose a suitable quantization scheme for such settings, referred to as universal vector quantization for FL (UVeQFed). We show that combining universal vector quantization methods with FL yields a decentralized training system in which the compression of the trained models induces only a minimum distortion. We then theoretically analyze the distortion, showing that it vanishes as the number of users grows. We also characterize how models trained with the conventional federated averaging method combined with UVeQFed  converge to the model which minimizes the loss function. Our numerical results demonstrate the gains of UVeQFed over previously proposed methods in terms of both  distortion induced in quantization and accuracy of the resulting aggregated model. In particular, we show that UVeQFed allows converging to a more accurate model when trained using the MNIST and CIFAR-10 data sets compared to existing schemes. 
	\end{abstract}
	
	\vspace{-0.4cm}
	\section{Introduction}
	\label{sec:Intro}
	\vspace{-0.1cm} 
	Machine learning methods have  demonstrated unprecedented performance in a broad range of applications \cite{lecun2015deep}. This is achieved by training a deep network model based on a large number of labeled training samples. Often, these samples are gathered on end devices or users, such as smartphones, while the deep model is maintained by a computationally powerful centralized  server \cite{chen2019deep}. Traditionally, the users  send their labeled data to the server, who in turn uses the massive amount of samples to train the model. However, data often contains private information, which the users may prefer not to share, and having each user transmit large volumes of training data to the server may induce a substantial load on the communication link. This gives rise to the need to adapt the network on the end-devices., i.e., train a centralized model in a distributed fashion~\cite{dean2012large}. 
	\ac{fl} proposed in \cite{mcmahan2016communication}, is a method to update such decentralized models. 
	Instead of requiring the users to share their possibly private labeled data, each user trains the network locally, and conveys its trained model updates to the server. The server then iteratively aggregates these updates into a global network \cite{bonawitz2019towards, kairouz2019advances}, commonly using some weighted average, also known as {\em federated averaging} \cite{mcmahan2016communication}. 
	
	One of the major challenges of \ac{fl} is the transfer of a large number of updated model parameters over the uplink communication channel from the users to the server, whose  throughput is typically constrained \cite{mcmahan2016communication,chen2019joint, li2019federated, kairouz2019advances}. 
	This challenge can be tackled by reducing the number of participating users, via, e.g., scheduling policies \cite{yang2019scheduling, amiri2020update}. An alternative strategy is to reduce the volume of data each user conveys, via sparsification or scalar quantization \cite{konevcny2016federated,lin2017deep, hardy2017distributed, aji2017sparse,wen2017terngrad, alistarh2017qsgd, horvath2019natural, reisizadeh2019fedpaq,horvath2019stochastic,bernstein2018signsgd}. 
	The work \cite{konevcny2016federated} proposed various methods for compressing the updates sent from the users to the server. These methods include random masks, subsampling, and probabilistic quantization.
	Sparsifying masks for compressing the gradients were proposed in \cite{lin2017deep, hardy2017distributed, aji2017sparse}.
	 Additional forms of probabilistic scalar quantization for \ac{fl} were considered in \cite{wen2017terngrad, alistarh2017qsgd, horvath2019natural, reisizadeh2019fedpaq,horvath2019stochastic}.  
	However, these approaches are suboptimal from a quantization theory perspective, as, e.g., discarding a random subset of the gradients can result in dominant distortion, while scalar quantization is  inferior to  vector quantization \cite[Ch. 23]{polyanskiy2014lecture}. This motivates the design and analysis of  quantization methods for facilitating updated model transfer in \ac{fl}, which minimize the error induced by quantization in the aggregated global model.
	
	Here, we design quantizers for distributed  training by tackling the uplink compression in \ac{fl} problem from a  quantization theory perspective.  \textcolor{NewColor}{We first discuss the requirements which one have to account for, and can possibly exploit, when designing quantization schemes for \ac{fl}.  We specifically identify that such quantization schemes are required to operate without knowing the distribution of the model updates, as such knowledge is unlikely to be available in \ac{fl}. We also note that the repeated communications between the server and users imply that they can share a source of local randomness, by, e.g., sharing a random seed, which can be utilized by the quantization mechanism.
}
	 Based on these properties, we propose a scheme following concepts from universal quantization \cite{zamir1992universal}, referred to as \ac{uveqfed}.  \ac{uveqfed} implements {\em subtractive dithered lattice quantization}, which is based on solid information theoretic arguments. \textcolor{NewColor}{In particular, such schemes are known to approach the most accurate achievable finite-bit representation, dictated by rate-distortion theory, to within a controllable gap  \cite{zamir1996lattice}, as well as achieve more accurate quantized representation compared to scalar quantization methods (probabilistic or deterministic) used in existing \ac{fl} works,   while meeting the aforementioned requirements. Consequently, \ac{uveqfed} allows \ac{fl} to operate reliably under strict bit rate constraints, due to its ability to reduce the distortion induced by the need to quantize the model updates, which results in more accurate learned models with faster convergence compared to previously proposed methods.}  
	
	
	We theoretically analyze the ability of the server to accurately recover the updated model when \ac{uveqfed} is employed. We show that the error induced by \ac{uveqfed} is mitigated by conventional federated averaging, and analyze the convergence of the global model to the one which minimize the loss function. Specifically, our analysis reveals that the resulting quantization error can be bounded by a term which vanishes as the number of users grows, regardless of the statistical model from which the data of each user is generated. This rigorously proves that the quantization distortion can be made arbitrarily small when a sufficient number of users contribute to the overall model. 
	Then, we study the convergence of \ac{sgd}-based federated averaging with \ac{uveqfed} in a statistically heterogeneous setup, where the training available at each user obeys a different distribution, as is commonly the case in \ac{fl} \cite{kairouz2019advances,li2019federated}. We prove that for strongly convex and smooth objectives, the expected  distance between the resulting \ac{fl} performance and the optimal one asymptotically decays as one over the number of iterations, which is the same order of convergence reported for \ac{fl} without communication constraints in heterogeneous setups \cite{li2019convergence}.
	 Finally, we show that these theoretical gains translate into \ac{fl} performance gains in a numerical study. We demonstrate that \ac{fl} with \ac{uveqfed} yields more accurate global models and faster convergence compared to previously proposed quantization approaches for such setups when operating under tight bit constraints of two and four bits per sample, considering  synthetic data as well as the MNIST and CIFAR-10 data sets.

	The rest of this paper is organized as follows: 
	Section~\ref{sec:Model}  presents the system model and identifies the requirements of \ac{fl} quantization.
	Section~\ref{sec:Quantization} details the proposed quantization system, and Section \ref{sec:Analysis} theoretically anaylzes its performance. 
	Experimental results are presented in Section~\ref{sec:Sims}.
	Section~\ref{sec:Conclusions}  concludes the paper.
	Proofs of the  results stated in the paper are detailed in the appendix.

	Throughout the paper, we use boldface lower-case letters for vectors, e.g., ${\myVec{x}}$;
	Matrices are denoted with boldface upper-case letters,  e.g., 
	$\myMat{M}$, where $\myMat{I}_n$ is the $n\times n$ identity matrix;   
	calligraphic letters, such as $\mySet{X}$, are used for sets.
	The $\ell_2$ norm and stochastic expectation are written as  $\|\cdot\|$ and $\E\{ \cdot \}$,  respectively. 
	%
	Finally, $\mySet{R}$ and $\mySet{Z}$ are the sets of real numbers and integers, respectively.


	
	\vspace{-0.2cm}
	\section{System Model}
	\label{sec:Model}
	\vspace{-0.1cm} 	
	In this section we detail the considered setup of \ac{fl} with bit-constrained model updates. To that aim, we first review  the conventional \ac{fl} setup in Section \ref{subsec:ModelFed}. Then, in Section \ref{subsec:ModelReq}, we formulate the problem and identify the unique requirements of quantizers utilized in \ac{fl} systems.  
	
	\vspace{-0.2cm}
	\subsection{Federated Learning}
	\label{subsec:ModelFed}
	\vspace{-0.1cm}
	We consider the conventional \ac{fl} framework proposed in 	 \cite{mcmahan2016communication}. Here, a centralized server is training a model consisting of $m$ parameters based on labeled  samples available at a set of $\Nusers$ remote users, in order to minimize some loss function $\ell(\cdot ; \cdot)$. Letting  $\{\myX_i\IdxK, \myY_i\IdxK \}_{i=1}^{\Ntraining_k}$ be the set of $\Ntraining_k$ labeled training samples available at the $k$th user, $k \in \{1,\ldots,\Nusers\} \triangleq \mySet{K}$, \ac{fl} aims at recovering the $m\times 1$ weights vector $\myWeights\Opt$ satisfying
	\begin{equation}
	\label{eqn:Objective}
	\myWeights\Opt = \mathop{\arg \min}\limits_{\myWeights}\left\{\Objective(\myWeights) \triangleq \sum_{k=1}^{\Nusers} \alpha_k\Objective_k(\myWeights) \right\}.
	\end{equation}
	Here, the weighting average coefficients $\{\alpha_k\}$ are non-negative satisfying $\sum\alpha_k = 1$, and the local objective functions are defined as the empirical average over the corresponding training set, i.e.,
	\begin{equation*}
	\Objective_k(\myWeights) \! \equiv \!  \Objective_k\big(\myWeights;\{\myX_i\IdxK, \myY_i\IdxK \}_{i=1}^{\Ntraining_k} \big) \! \triangleq \! \frac{1}{\Ntraining_k}\sum_{i=1}^{\Ntraining_k}\ell\big(\myWeights;(\myX_i\IdxK, \myY_i\IdxK ) \big).
	\end{equation*}
	
	{\em Federated averaging} \cite{mcmahan2016communication} aims at recovering  $\myWeights\Opt$ using iterative subsequent updates. In each update of time instance $t$, the server shares its current model, represented by the vector $\myWeights_t \in \mySet{R}^{m}$, with the users. The $k$th user, $k \in  \mySet{K}$, uses its set of $\Ntraining_k$ labeled training samples to retrain the model $\myWeights_t$ over $\SGDIter$ time instances into an updated model $\myGlobalModel_{t+\SGDIter}\IdxK \in \mySet{R}^{m}$. 
	
	Having updated the model weights, the $k$th user should convey its model update,  
	\begin{figure}
		\centering
		{\includefig{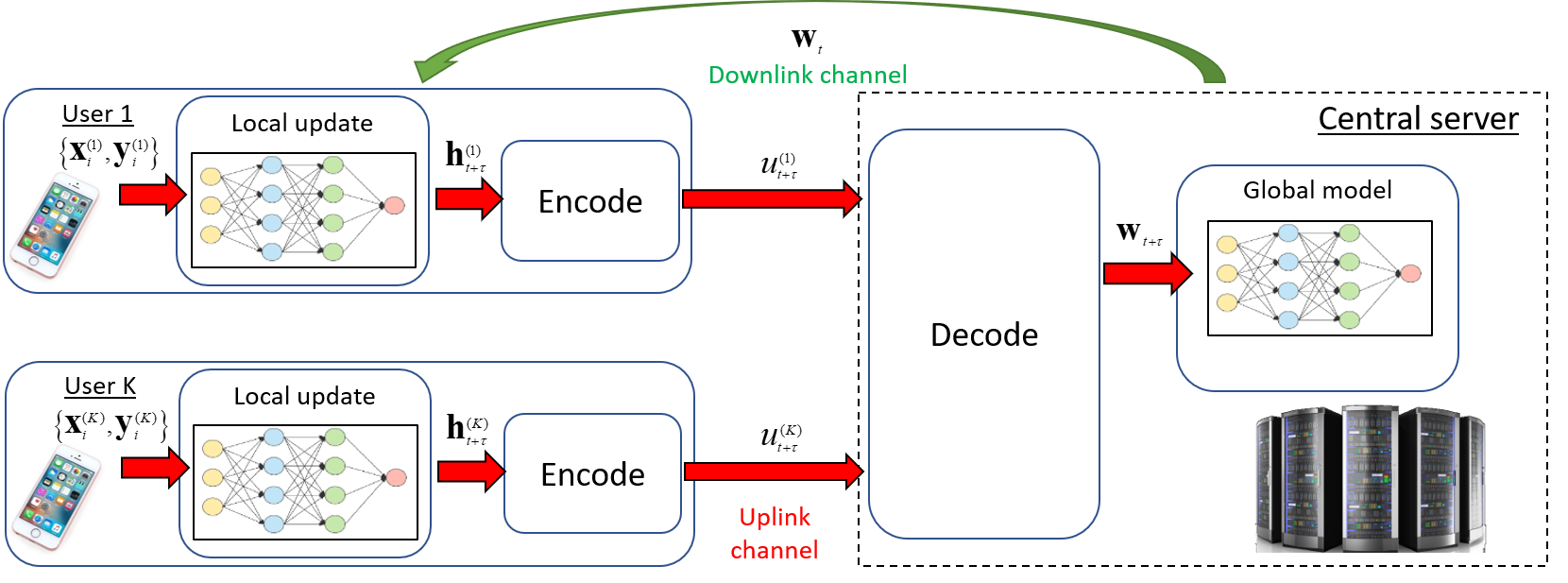}} 
		\caption{Federated learning with bit rate constraints.}
		\label{fig:GenSetup2}	 
	\end{figure} 
	denoted as $\myUpdate_{t+\SGDIter}\IdxK \triangleq \myGlobalModel_{t+\SGDIter}\IdxK - \myWeights_t$, to the server. Since uploading throughput is typically more limited compared to its downloading counterpart \cite{speedtest2019}, the $k$th user needs to communicate a finite-bit quantized representation of its model update.
	Quantization consists of encoding the model update into a set of bits, and decoding each  bit combination into a recovered model update \cite{gray1998quantization}.  The $k$th model update $\myUpdate_{t+\SGDIter}\IdxK$ is therefore encoded into a digital codeword of $\Rate_k$ bits denoted as $u_t\IdxK \in \{0,\ldots,2^{\Rate_k}-1\} \triangleq \mySet{U}_k$, using an encoding function whose input is $\myUpdate_{t+\SGDIter}\IdxK$, i.e., 	 
	\begin{equation}
	\label{eqn:Encoder}
	e_{t+\SGDIter}\IdxK:\mySet{R}^{m} \mapsto \mySet{U}_k. 
	\end{equation}
	
	The uplink channel is modeled as a bit-constrained link,  as commonly assumed in the \ac{fl} literature  \cite{konevcny2016federated,lin2017deep, hardy2017distributed, aji2017sparse,wen2017terngrad, alistarh2017qsgd, horvath2019natural, reisizadeh2019fedpaq,horvath2019stochastic,bernstein2018signsgd}. \textcolor{NewColor}{In such channels, each $R_k$ bit codeword is recovered by the server without errors, representing, e.g., coded communications at rates below the channel capacity, where arbitrarily small error rates can be guaranteed by proper channel coding. }
	The server uses the received codewords $\{u_{t+\SGDIter}\IdxK\}_{k=1}^{\Nusers}$ to reconstruct $\hat{\myUpdate}_{t+\SGDIter} \in  \mySet{R}^{m}$, obtained via a joint decoding function  
	\begin{equation}
	\label{eqn:decoder}
	d_{t+\SGDIter}: \mySet{U}_1\times \ldots \times\mySet{U}_\Nusers   \mapsto \mySet{R}^{m}. 
	\end{equation}	 
	The recovered $\hat{\myUpdate}_{t+\SGDIter}$ is an estimate of  the  weighted average $\sum_{k=1}^{\Nusers}\alpha_k\myUpdate_{t+\SGDIter}\IdxK$. Finally, the global model $\myWeights_{t+\SGDIter}$ is updated via 
	\begin{equation}
	\label{eqn:GlobalModel}
	\myWeights_{t+\SGDIter} = \myWeights_t + \hat{\myUpdate}_{t+\SGDIter}. 
	\end{equation}   
	An illustration of this \ac{fl} procedure is depicted in Fig. \ref{fig:GenSetup2}.
%
	%
	%
	Clearly, if the number of allowed bits is sufficiently large,   the distance $ \| \hat{\myUpdate}_{t+\SGDIter} -\sum_{k=1}^{\Nusers}\alpha_k\myUpdate_{t+\SGDIter}\IdxK\|^2$ can be made arbitrarily small, allowing the server to update the global model as the desired weighted average, \textcolor{NewColor}{denoted $\myWeights_{t+\SGDIter}\Desired$,} via:  
	\begin{equation}
	\label{eqn:UpdatedModel}
	\myWeights_{t+\SGDIter}\Desired = \sum_{k=1}^{\Nusers}\alpha_k \myGlobalModel_{t+\SGDIter}\IdxK. 
	\vspace{0.1cm}
	\end{equation}  

	In the presence of a limited bit budget, i.e., small values of $\{\Rate_k\}$, distortion is induced which can severely degrade the ability of the server to  update its model. 
	To tackle this issue,  various methods have been proposed for quantizing the model updates, commonly based on sparsification or probabilistic scalar quantization. 	
	These approaches are  suboptimal from a quantization theory perspective, \textcolor{NewColor}{namely, the gap between the distortion they achieve in quantizing a signal using a given number of bit and the most accurate achievable finite-bit representation,
	dictated by rate-distortion theory, can be further reduced by, e.g., using  vector quantization \cite[Ch. 23]{polyanskiy2014lecture}.} This motivates the study of efficient and practical quantization methods for \ac{fl}. 

	\vspace{-0.2cm}
	\subsection{Problem Formulation}
	\label{subsec:ModelReq}
	\vspace{-0.1cm}
	Our goal is to propose an encoding-decoding system which mitigates the effect of quantization errors on the ability of the server to accurately recover the updated model \eqref{eqn:UpdatedModel}. 
	To faithfully represent the \ac{fl} setup, we design our quantization strategy in light of the following requirements and assumptions:
	\begin{enumerate}[label={\em A\arabic*}]
		\item \label{itm:As2} All users share the same encoding function, denoted as $e_t\IdxK(\cdot) = e_t(\cdot)$ for each $k \in \mySet{K}$. This requirement, which was also considered in \cite{konevcny2016federated}, significantly simplifies \ac{fl}  implementation.
		\item \label{itm:As3} {\em a-priori} knowledge or distribution of $\myUpdate_{t+\SGDIter}\IdxK$ is assumed.  
		\item \label{itm:As4} As in \cite{konevcny2016federated}, the users and the server share a source of common randomness. This is achieved by, e.g., letting the server share with each user a random seed along with the weights. Once a different seed is conveyed to each user, it can be used to obtain a dedicated source of common randomness shared by server and each of the users 		
		for the entire \ac{fl} procedure. 
	\end{enumerate}

	Requirement \ref{itm:As3} gives rise to the need for a {\em universal quantization} approach, namely, a scheme which operates reliably regardless of the distribution of the model updates and without its prior knowledge.
	 In light of the above requirements, we propose \ac{uveqfed} in the following section.

	\vspace{-0.2cm}
	\section{\ac{uveqfed}}
	\label{sec:Quantization}
	\vspace{-0.1cm} 
	We now propose \ac{uveqfed}, which conveys the model updates  
	 $\{\myUpdate_{t+\SGDIter}\IdxK\}$ from the users to the server over the rate-constrained channel  
	using a universal quantization method. 
	Specifically, the scheme encodes each model update using {\em subtractive dithered lattice quantization} \cite{zamir1992universal}, which operates in the same manner for each user, satisfying \ref{itm:As2}. \ac{uveqfed} allows the server to recover the updates with small average error regardless of the distribution of $\{\myUpdate_{t+\SGDIter}\IdxK\}$, as required in \ref{itm:As3}, by exploiting the source of common randomness assumed in \ref{itm:As4}.  
In addition to its compliance with the model requirements stated in Section~\ref{subsec:ModelReq}, the proposed approach is particularly suitable for \ac{fl}, as the distortion is mitigated by federated averaging, as we prove in Section~\ref{sec:Analysis}.	
	This significantly improves the overall \ac{fl} capabilities, as numerically demonstrated in Section~\ref{sec:Sims}. 
	The proposed quantization method is detailed in Section~\ref{subsec:QuantScheme}, followed by a  discussion in Section~\ref{subsec:QuantDiscussion}.

	\vspace{-0.2cm}
	\subsection{Quantization Scheme}
	\label{subsec:QuantScheme}
	\vspace{-0.1cm} 
	Here, we present  the encoding and decoding functions, $e_{t+\SGDIter}(\cdot)$ and $d_{t+\SGDIter}(\cdot)$. 
	Following requirement \ref{itm:As2}, we utilize  universal vector quantization, i.e., a quantization scheme which maps each set of continuous-amplitude values into a discrete representation in a manner which is ignorant of the underlying distribution. Common universal quantization methods are based on selection from an ensemble of source codes \cite{chou1996vector}, or alternatively, on subtractive dithering \cite{ziv1985universal,gray1993dithered,lipshitz1992quantization,zamir1992universal,zamir1999multiterminal,kirac1996results}, where the latter is simpler to implement being based on adding dither, i.e., noise, to the discretized quantity, but requires knowledge of the dither as it is subtracted from the discrete quantity when parsing the quantized value. The source of common randomness assumed in \ref{itm:As4} implies that the server and the users can generate the same realizations of a dither signal. We thus design \ac{uveqfed} based on dithered vector quantization, and particularly, on lattice quantization, detailed in the following.
		
	Let $\Nlattice$ be a fixed positive integer, referred to henceforth as the lattice dimension, and a let $\GenMat$ be a non-singular $\Nlattice \times \Nlattice$ matrix, which denotes the lattice generator matrix. 
	For simplicity, we assume that $\Npoints \triangleq \frac{ m} {\Nlattice}$ is an integer, where $m$ is the number of model parameters, although the scheme can also be applied when this does not hold by replacing $\Npoints$ with $\lceil \Npoints \rceil$. 
	Next, we use $\mySet{L}$ to denote the lattice, which is the set of points in $\mySet{R}^{\Nlattice}$ that can be written as an integer linear combination of the columns of $\GenMat$, i.e.,  \textcolor{NewColor}{the set of all points  $ \myVec{x} \in \mySet{R}^{\Nlattice}$ which can be written as $\GenMat \myVec{l}$ with $\myVec{l}$ having integer entries:}
	\begin{equation}
	\label{eqn:Lattice}
	\mySet{L} \triangleq \{ \myVec{x}  = \GenMat \myVec{l}: \myVec{l} \in \mySet{Z}^{\Nlattice} \}. 
	\end{equation} 
	
	A lattice quantizer $Q_\mySet{L}(\cdot)$ maps each  $ \myVec{x} \in \mySet{R}^{\Nlattice}$ to its nearest lattice point, i.e., $Q_\mySet{L}\left( \myVec{x}\right) = \myVec{l}_{\myVec{x}}$ where $ \myVec{l}_{\myVec{x}} \in \mySet{L}$ if  $\|\myVec{x} - \myVec{l}_{\myVec{x}}\| \le \|\myVec{x} - \myVec{l}\|$ for every $\myVec{l} \in \mySet{L}$.  
	Finally, let  $\Partition$ be the basic lattice cell  \cite{zamir1996lattice}, i.e., the set of points  $\myVec{x} \in \mySet{R}^{\Nlattice}$ which are closer to $\myVec{0}$ than to any other lattice  point:  
	\begin{equation}
	\label{eqn:BasicCell}
	\Partition \triangleq \{\myVec{x} \in \mySet{R}^{\Nlattice} : \|\myVec{x}\|  <  \|\myVec{x} - \myVec{p}\|, \forall \myVec{p} \in \mySet{L} / \{\myVec{0}\}  \}. 
	\end{equation} 	
	\textcolor{NewColor}{As $\Partition$ represents the set of points  which are closer to the origin than to any other lattice point, its shape depends on the lattice $\mySet{L}$, and in particular on the generator matrix $\GenMat$. For instance, when $\GenMat = \Delta \cdot \myI_{\Nlattice}$ for some $\Delta > 0$, then $\mySet{L}$ is the square lattice, for which $\Partition$ is the set of vectors $\myVec{x} \in \mySet{R}^{\Nlattice} $ whose $\ell_\infty$ norm is not larger than $\frac{\Delta}{2}$. In the two-dimensional case, such a generator matrix results in  $\Partition$ being a square centered at the origin.  For this setting,  $Q_{\mySet{L}}(\cdot)$ implements entry-wise scalar uniform quantization with spacing $\Delta$ \cite[Ch. 23]{polyanskiy2014lecture}. In general, the basic cell can take different shapes, such as hexagons for two-dimensional hexagonal lattices.}

	Using the above  definitions in lattice quantization, we now present the encoding and decoding procedures of \ac{uveqfed}, which are based on subtractive dithered lattice quantization: 
	
{\bf Encoder:}	\textcolor{NewColor}{The proposed encoding function $e_{t+\SGDIter}(\cdot)$ implements dithered lattice quantization in four stages. It first normalizes the model updates and partitions it into sub-vectors of the lattice dimension, where the normalization is used to prevent overloading the finite lattice. Then, each vector is dithered before it is quantized to result in a distortion term which is not deterministically determined by the model updates, and is thus reduced by averaging. The quantized representation is  compressed in  lossless manner using entropy coding to further reduce its volume without inducing additional distortion.} These steps are detailed in the following:
	\begin{enumerate}[label={\em E\arabic*}]
		\item \label{itm:Partition} {\bf Normalize and partition:} The $k$th user scales $\myUpdate_{t+\SGDIter}\IdxK$ by $\ScaleFactor\|\myUpdate_{t+\SGDIter}\IdxK\|$ for some $\ScaleFactor >0$, and divides the result into $\Npoints$ distinct $\Nlattice \times 1$ vectors, denoted $\{\myUpdateV_i\IdxK \}_{i=1}^{\Npoints}$. The scalar quantity $\ScaleFactor\|\myUpdate_{t+\SGDIter}\IdxK\|$ is quantized separately from $\{\myUpdateV_i\IdxK \}_{i=1}^{\Npoints}$ using some fine-resolution quantizer.
		\item \label{itm:Dither}  {\bf Dithering:} The encoder utilizes the source of common randomness, e.g., a shared seed, to generate the set of  $\Nlattice \times 1$ dither vectors $\{\myVec{z}_i\IdxK \}_{i=1}^{\Npoints}$, which are randomized in an i.i.d. fashion, independently of $\myUpdate\IdxK_{t+\SGDIter}$, from a uniform distribution over $\Partition$.  
		\item \label{itm:Quantize} {\bf Quantization:} The vectors  $\{\myUpdateV_i\IdxK \}_{i=1}^{\Npoints}$ are discretized by adding the dither vectors and applying lattice quantization, i.e., by computing $\{Q_\mySet{L}(\myUpdateV_i\IdxK +\myVec{z}_i\IdxK  )\}$.
		\item \label{itm:Coding} {\bf Entropy coding:} The discrete values $\{Q_\mySet{L}(\myUpdateV_i\IdxK +\myVec{z}_i\IdxK  )\}$ are encoded into a digital codeword $u_{t+\SGDIter}\IdxK$ in a lossless manner.   
	\end{enumerate}

	In order to utilize entropy coding in step \ref{itm:Coding}, the discretized  $\{Q_\mySet{L}(\myUpdateV_i\IdxK +\myVec{z}_i\IdxK  )\}$ must take values on a {\em finite set}. This is achieved by the normalization in Step \ref{itm:Partition}, which guarantees that   $\{\myUpdateV_i\IdxK \}_{i=1}^{\Npoints}$ all reside inside the $\Nlattice$-dimensional ball with radius $\ScaleFactor^{-1}$, in which the number of lattice points is not larger than $\frac{\pi^{\Nlattice/2}}{\ScaleFactor^{\Nlattice}\Gamma(1+\Nlattice/2)\det(\GenMat)}$ \cite[Ch. 2]{conway2013sphere}, where $\Gamma(\cdot)$ is the Gamma function.  The overhead in accurately quantizing the single scalar quantity $\ScaleFactor\|\myUpdate\IdxK\|$ is typically negligible compared to the number of bits required to convey the set of vectors $\{\myUpdateV_i\IdxK \}_{i=1}^{\Npoints}$,  hardly affecting the overall quantization rate.

{\bf Decoder:}	\textcolor{NewColor}{The decoding mapping $d_{t+\SGDIter}(\cdot)$ is comprised of four stages. The purpose of the first three steps is to invert the encoding procedure by decoding the lossless entropy code used in \ref{itm:Coding}, subtracting the dither added in \ref{itm:Dither}, and reforming the full model update vector from the partitioned sub-vectors generated in \ref{itm:Partition}. The final stage uses the recovered model update to compute the aggregated global model.} These stages are detailed in the following:
	\begin{enumerate}[label={\em D\arabic*}]
		\item \label{itm:Decoding} {\bf Entropy decoding:} The server first decodes each digital codeword $u_{t+\SGDIter}\IdxK$ into the discrete value $\{Q_\mySet{L}(\myUpdateV_i\IdxK +\myVec{z}_i\IdxK  )\}$. Since the encoding is carried out using a lossless source code, the discrete values are recovered without any errors.  
		\item \label{itm:SubDither} {\bf Dither subtraction:} Using the source of common randomness, the server generates the dither vectors $\{\myVec{z}_i\IdxK \}$, which can be carried out rapidly and at low complexity using  random number generators as the dither vectors obey a uniform distribution. The server then subtracts the corresponding vector from each lattice point, i.e., compute  $\{Q_\mySet{L}(\myUpdateV_i\IdxK +\myVec{z}_i\IdxK  )- \myVec{z}_i\IdxK\} $. An illustration of the subtractive dithered lattice quantization procedure is illustrated in Fig. \ref{fig:DithQuan1}. 
		\begin{figure}
			\centering
			{\includegraphics[width=\figWidth]{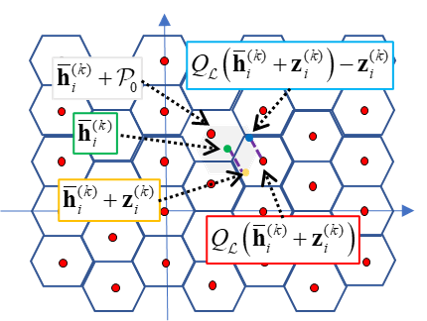}} 
			\caption{Subtractive dithered lattice quantization illustration.}
			\label{fig:DithQuan1}	 
		\end{figure}
		\item \label{itm:Collect} {\bf Collecting and scaling:} The values  $\{Q_\mySet{L}(\myVec{h}_i\IdxK +\myVec{z}_i\IdxK  )- \myVec{z}_i\IdxK\} $ are collected into an $m \times 1$ vector $\hat{\myUpdate}\IdxK_{t+\SGDIter}$ using the inverse operation of the partitioning and normalization in Step \ref{itm:Partition}.
		\item \label{itm:Recovery}  {\bf Model recovery:} The recovered matrices are combined into an updated model based on \eqref{eqn:GlobalModel}.  
		Namely,
			\begin{equation}
			\label{eqn:ModelUpdate1}
		\myWeights_{t+\SGDIter} = \myWeights_t +\sum_{k=1}^{\Nusers}\alpha_k\hat{\myUpdate}_{t+\SGDIter}\IdxK.
			\end{equation}	  
	\end{enumerate}

\begin{figure*}
		\centering
		{\includegraphics[width = 0.9\linewidth]{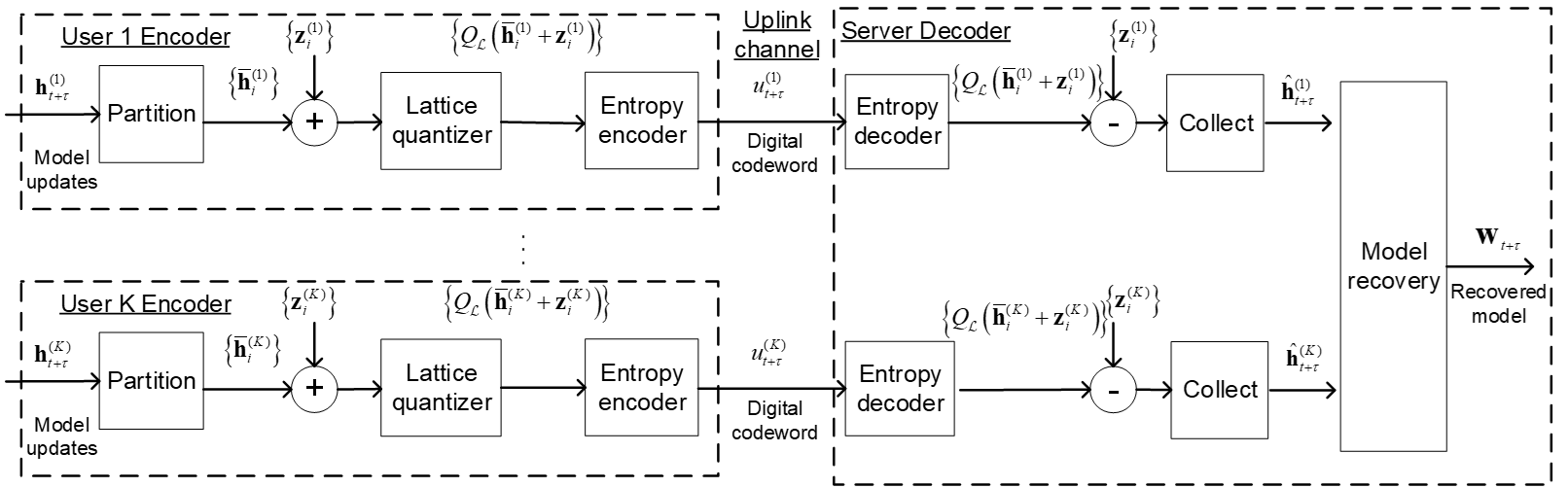}} 
		\caption{\ac{uveqfed} encoding-decoding block diagram.} 
		\label{fig:universalsystem1}	 
	\end{figure*}

	A block diagram of the proposed scheme is depicted in Fig.~\ref{fig:universalsystem1}.  The usage of subtractive dithered lattice quantization in Steps \ref{itm:Dither}-\ref{itm:Quantize} and \ref{itm:SubDither} allow obtaining a digital representation which is relatively close to the true quantity, as illustrated in Fig. \ref{fig:DithQuan1}, without relying on prior knowledge of its distribution.	The joint decoding aspect of the proposed scheme is introduced in the final model recovery Step \ref{itm:Recovery}. The remaining encoding-decoding procedure, i.e., Steps \ref{itm:Partition}-\ref{itm:Collect} is carried out independently for each user. 
%

	

	\vspace{-0.2cm}
	\subsection{Discussion}
	\label{subsec:QuantDiscussion}
	\vspace{-0.1cm}	 
	\ac{uveqfed} has several clear advantages. While it is based on information theoretic arguments, the resulting architecture is rather simple to implement. \textcolor{NewColor}{Both subtractive dithered quantization as well as entropy coding are concrete and established methods which can be realized with relatively low complexity and feasible hardware requirements. In particular, the main novel aspect of \ac{uveqfed}, i.e., the usage of subtractive dithered lattice quantization, first requires generating the dither signal via, e.g., the methods discussed in \cite{rubinstein1982generating} for randomizing uniformly distributed random vectors. Then, the encoder carries out lattice projection of each sub-vector which, for finite and small $\Nlattice$ as used in the numerical study in Section~\ref{sec:Sims},  involves a complexity term which only grows linearly with the number of parameters $m$. This resulting additional complexity is of the same order as previous quantized \ac{fl} strategies, e.g., QSGD \cite{alistarh2017qsgd} which also uses entropy coding, and is typically dominated by the computational burden involved in training a deep model with $m$ parameters.} 
	The source of common randomness needed for generating the dither vectors can be obtained by sharing a common seed between the server and users, as also assumed in \cite{konevcny2016federated}.
	The statistical characterization of the quantization error of such quantizers does not depend on the distribution of the model updates. This analytical tractability allows us to rigorously show that its combination with federated averaging mitigates the quantization error in Section \ref{sec:Analysis}.  A similar approach was also used in the analysis of probabilistic quantization schemes for average consensus problems \cite{aysal2008distributed}. 
	
	As the updates under the considered system model are quantized for a specific task, i.e., to obtain the global model by  averaging, \ac{fl} with bit constraints can be treated as a task-based quantization scenario \cite{shlezinger2020task,shlezinger2018hardware,shlezinger2018asymptotic,Salamtian19task}. In \ac{uveqfed}, this task is accounted for in the selection of the quantization scheme, using one for which the distortion vanishes by averaging regardless of the values of $\{\myUpdate\IdxK\}$. 
	In addition, \ac{uveqfed} is derived based on modeling the uplink channel as a bit-limited pipeline. While this model is widely adopted in the \ac{fl} literature, it may not accurately reflect the true nature of wireless communication channels, being a noisy and shared media. \textcolor{NewColor}{This property of wireless communications is known to affect the design of deep learning systems operating over such channels \cite{yang2020federated, sery2020over, letaief2019roadmap, kang2019incentive, shlezinger2020communication, kang2020reliable, chen2019artificial}}. Traditionally, transmitted bit streams such as the codewords produced by \ac{uveqfed} are protected using a separate channel code for mitigating the errors induced by the noisy channel. Alternatively, one can also consider extending \ac{uveqfed} to directly map the model updates into a channel codeword as a form of task-based joint source channel coding. We leave this extension of \ac{uveqfed} for future work.
	
	The encoding Steps \ref{itm:Partition}-\ref{itm:Quantize} can be viewed as a generalization of probabilistic scalar quantizers, used in, e.g., QSGD \cite{alistarh2017qsgd}. When the lattice dimension is $\Nlattice = 1$ and   $\ScaleFactor=1$, Steps \ref{itm:Partition}-\ref{itm:Quantize} implement the same encoding as QSGD. However, the decoder is not the same as in QSGD due to the dither subtraction in Step \ref{itm:SubDither}, which is known to reduce the distortion  and yield an error term that does not depend on the model updates \cite{gray1993dithered}. Furthermore, \ac{uveqfed} allows using vector quantizers, i.e., setting $\Nlattice > 1$, which is known to further improve the quantization accuracy \cite{zamir1996lattice}. Specifically, the usage of vector quantizers allows \ac{uveqfed} to combine dimensionality reduction methods with quantization schemes by jointly mapping sets of samples into discrete representations. The gains of subtracting the dither at the decoder and of using vector quantizers over scalar ones are numerically demonstrated in our experimental study in Section \ref{sec:Sims}.
		
	
	The usage of lossless source coding in Steps \ref{itm:Coding} and \ref{itm:Decoding} allows exploiting the typically non-uniform distribution of the quantizer outputs. A similar approach was also used in QSGD \cite{alistarh2017qsgd}, where Elias codes were utilized.
	Since  Steps \ref{itm:Coding} and \ref{itm:Decoding} involve multiple encoders and a single decoder, improved compression can be achieved by utilizing distributed source coding methods, e.g., Slepian-Wolf coding \cite[Ch. 15.4]{cover2012elements}. In such cases, the server decodes the received codewords  
	$\{u_{t+\SGDIter}\IdxK\}$ into $\{Q_\mySet{L}(\myUpdateV_i\IdxK +\myVec{z}_i\IdxK  )\}$  
	in  a joint manner, instead of decoding each $Q_\mySet{L}(\myUpdateV_i\IdxK +\myVec{z}_i\IdxK  )$ from its corresponding $u_{t+\SGDIter}\IdxK$ separately. 
	Similarly, the distributed nature of \ac{fl} can be exploited to optimize the reconstruction fideitly for a given bit budget using Wyner-Ziv coding \cite{wyner1976rate}.
	However, such distributed coding schemes typically require a-priori knowledge of the joint distribution of $\{\myUpdateV_i\IdxK\}$, and utilize different encoding mappings for each user, thus not meeting  requirements \ref{itm:As2}-\ref{itm:As3}. 
	
%
	
	Finally, we note that the \ac{fl} performance is affected by the selection of the lattice $\mySet{L}$ and the  coefficient $\ScaleFactor$. In general, lattices of higher dimensions typically result in more accurate representations, at the cost of increased complexity. Methods for designing the lattice generator matrix $\GenMat$ can be found in  \cite{agrell1998optimization}. The coefficient $\ScaleFactor$ should be set to allow the usage of a limited number of lattice points, which is translated into less bits, without concentrating the resulting vectors such that they become indistinguishable after quantization. For example, using $\ScaleFactor = 1$ results in most quantized values mapped to zero, as also observed in \cite{alistarh2017qsgd}. A reasonable setting is $\ScaleFactor = 3\frac{1}{\sqrt{M}}$, resulting in $\ScaleFactor\|\myUpdate\IdxK\|$ approaching $3$ times the standard deviation of the quantized vectors when they are zero-mean and i.i.d., and thus assuring that they reside inside the unit $\Nlattice$-ball with probability of over $88\%$ by Chebyshev's inequality \cite{ferentios1982tcebycheff}.

	\vspace{-0.2cm}
	\section{Performance Analysis}
	\label{sec:Analysis}
	\vspace{-0.1cm} 
	Next, we analyze the performance of \ac{uveqfed}, characterizing its distortion and studying its convergence properties.   We consider the conventional local \ac{sgd} training method, detailed in Section \ref{subsec:QuantLocal}, and characterize the resulting distortion of \ac{uveqfed} and the convergence of the global model in Sections \ref{subsec:QuantAnalysis}-\ref{subsec:ConvergenceAnalysis}, respectively. 
	
	\vspace{-0.2cm}
	\subsection{Local SGD}
	\label{subsec:QuantLocal}
	\vspace{-0.1cm}
	Local \ac{sgd} is arguably the most common training method used for federated averaging \cite{stich2018local}. Here, each user updates the weights using $\SGDIter$ \ac{sgd} iterations before sending the updated model to the server for aggregation. 	
	 Let $\Qerror_{t}^{(k)}$ denote the error induced in quantizing the model update $\myUpdate_{t}\IdxK$, and let $i_t\IdxK$ be the sample index chosen uniformly from the local data of the $k$th user at time  $t$.  
	 By defining the gradient computed at a single sample of index $i$ as $ \nabla\Objective_{k}^i(\myGlobalModel) \triangleq  \nabla\Objective_{k}\big(\myGlobalModel;(\myX_{i}\IdxK, \myY_{i}\IdxK ) \big)$, the  local weights at the $k$th user, denoted $\myGlobalModel_t\IdxK$, are updated via:
	 \ifsingle
	 \begin{align} 
	 & \myGlobalModel_{t+1}\IdxK \!= \! 
	 \begin{cases}
	 \myGlobalModel_{t}\IdxK - \StepSize_t \nabla\Objective_k^{i_t\IdxK}\big(\myGlobalModel_{t}\IdxK  \big), & \!\!t\!+\!1 \!\notin \!\mySet{T}_{\SGDIter}, \\
	 \sum\limits_{k'=1}^{\Nusers}\!\alpha_{k'}\! \Big(\! \myGlobalModel_{t}\IdxKt \!\!-\! \StepSize_t \nabla\Objective_{k'}^{i_t\IdxKt}\!\!\big(\myGlobalModel_{t}\IdxKt \big) \!+\! \Qerror_{t+1}\IdxKt\Big), & \!\! t\!+\!1 \!\in\! \mySet{T}_{\SGDIter},
	 \end{cases} 
	 \label{eqn:SGD1}
	 \end{align}
	 \else
	 \begin{align} 
	& \myGlobalModel_{t+1}\IdxK \!= \! \notag \\
	 &\begin{cases}
	 \myGlobalModel_{t}\IdxK - \StepSize_t \nabla\Objective_k^{i_t\IdxK}\big(\myGlobalModel_{t}\IdxK  \big), & \!\!t\!+\!1 \!\notin \!\mySet{T}_{\SGDIter}, \\
	 \sum\limits_{k'=1}^{\Nusers}\!\alpha_{k'}\! \left( \myGlobalModel_{t}\IdxKt \!\!-\! \StepSize_t \nabla\Objective_{k'}^{i_t\IdxKt}\!\!\big(\myGlobalModel_{t}\IdxKt \big) \!+\! \Qerror_{t+1}\IdxKt\right), & \!\! t\!+\!1 \!\in\! \mySet{T}_{\SGDIter},
	 \end{cases} 
	 \label{eqn:SGD1}
	 \end{align}
	 \fi
	 where $\StepSize_t$ is the learning rate at time instance $t$, and $\mySet{T}_{\SGDIter}$ is the set of positive integer multiples of $\SGDIter$.  
	 
	 We focus on the case in which the users compute a single stochastic gradient in each time instance. Hence, the performance in terms of convergence rate can be further improved by using mini-batches \cite{stich2018local}, i.e., replacing the random index   $i_t\IdxK$ with a set of random indices.
	 The fact that the model updates are quantized when conveyed to the server is encapsulated in the per-user model update quantization error $\Qerror_{t}\IdxK$.

	\vspace{-0.2cm}
	\subsection{Quantization Error Bound}
	\label{subsec:QuantAnalysis}
	\vspace{-0.1cm} 
	The need to represent the model updates $\myUpdate_{t+\SGDIter}\IdxK$ using a finite number of bits inherently induces some distortion, i.e., the recovered vector is $\hat{\myUpdate}_{t+\SGDIter}\IdxK = \myUpdate_{t+\SGDIter}\IdxK + \Qerror_{t+\SGDIter}\IdxK$. 
	The error in representing $\ScaleFactor\|\myUpdate_{t+\SGDIter}\IdxK\|$ is assumed to be negligible. For example,  the normalized quantization error is of the order of $10^{-7}$ for $12$ bit quantization of a scalar value, and decreases exponentially with each additional bit \cite[Ch. 23]{polyanskiy2014lecture}. 
	 Letting $\LatFactor$ be the normalized second order lattice moment, defined as  
	$\LatFactor \triangleq {\int_{\Partition} \|\myVec{x}\|^2 d\myVec{x}}/{\int_{\Partition}  d\myVec{x}}$ \cite{conway1982voronoi}, the moments of the quantization error satisfy the following:
	\begin{theorem}
		\label{thm:QError}
		The quantization error vector $\Qerror_{t+\SGDIter}\IdxK$  has zero-mean entries and satisfies
		\begin{equation}
		\label{eqn:QError}
		\E\big\{\big\|\Qerror_{t+\SGDIter}\IdxK \big\|^2\big| \myUpdate_{t+\SGDIter}\IdxK \big\} =\ScaleFactor^2\|\myUpdate_{t+\SGDIter}\IdxK\|^2 M \LatFactor.
		\end{equation}		 
	\end{theorem}
	 
	 \ifFullVersion	
	 {\em Proof:}
	 See Appendix \ref{app:Proof0}.
	 \fi
	 
	 \smallskip

	
	Theorem \ref{thm:QError} characterizes the distortion in quantizing the model updates using \ac{uveqfed}. Unlike the corresponding characterization of previous quantizers used in \ac{fl} which obtained an upper bound on the quantization error, e.g., \cite[Lem. 1]{alistarh2017qsgd},   the dependence of the expected error norm on the number of bits is not explicit in \eqref{eqn:QError}, but rather encapsulated in the lattice moment $\LatFactor$. To observe that \eqref{eqn:QError} indeed represents lower distortion compared to previous \ac{fl} quantization schemes, we note that even when scalar quantizers are used, i.e., $\Nlattice = 1$ for which $\frac{1}{\Nlattice}\LatFactor$ is known to be largest \cite{conway1982voronoi}, the resulting quantization is reduced by a factor of $2$ compared to conventional probabilistic scalar quantizers, such as QSGD, due to the subtraction of the dither upon decoding in Step \ref{itm:SubDither} \cite[Thms. 1-2]{gray1993dithered}.   
	
	The model updates are recovered in order to update the global model via $\myWeights_{t+\SGDIter} = \sum\alpha_{k}\myGlobalModel_{t+\SGDIter}\IdxK$ at the server. We next show that the statistical characterization of the distortion in Theorem~\ref{thm:QError} contributes to the accuracy in recovering the desired  $\myWeights_{t+\SGDIter}\Desired$ \eqref{eqn:UpdatedModel} via $\myWeights_{t+\SGDIter}$. To that aim, we introduce the following assumption on the stochastic gradients, which is often employed in distributed learning studies  \cite{stich2018local,li2019convergence,zhang2013communication}:
		\begin{enumerate}[label={\em AS\arabic*}] 
		\item \label{itm:C3} The expected squared $\ell_2$ norm of the random vector  $\nabla\Objective_k^{i}\big(\myWeights  \big)$, representing the stochastic gradient evaluated at $\myWeights$, is bounded by some $\GradMom_k > 0$  for all $\myWeights \in \mySet{R}^{m}$.   
	\end{enumerate}

	We can now bound the distance between the desired model $\myWeights_{t+\SGDIter}\Desired$ and the recovered one $\myWeights_{t+\SGDIter}$, as stated in the following theorem:  
	\begin{theorem}
		\label{thm:DistBound} 
		When \ref{itm:C3} holds, the mean-squared distance between $\myWeights_{t+\SGDIter}$ and    $\myWeights_{t+\SGDIter}\Desired$ satisfies
		\vspace{-0.1cm}
		\begin{equation}
		\label{eqn:DistBound}
		\!	\E\left\{ \left\|\myWeights_{t\!+\!\SGDIter}\! -\! \myWeights_{t\!+\!\SGDIter}\Desired\right\|^2   \right\} \!\leq\!  M\ScaleFactor^2 \LatFactor \SGDIter \!\left( \sum_{t'=t}^{t\!+\!\SGDIter \!- \!1} \StepSize_{t'}^2\right)\!  \sum\limits_{k=1}^{\Nusers} \alpha_k^2\GradMom_k.
		\end{equation}
	\end{theorem}
	{\em Proof:}
	See Appendix \ref{app:Proof1}. 
	
	\smallskip
	
	Theorem \ref{thm:DistBound} implies that the recovered model can be made arbitrarily close to the desired one by increasing $\Nusers$, namely, the number of users. 
	For example, when $\alpha_k =  1/\Nusers$, i.e., conventional averaging, it follows from Theorem \ref{thm:DistBound} that the   mean-squared error in the weights decreases as $1/\Nusers$. 
	In particular, if $\mathop{\max}_{k   }  \alpha_k$ decreases with $\Nusers$, which  essentially means that the updated model is not based only on a small part of the participating users, then the distortion  vanishes in the aggregation process. Furthermore, when the step size $\StepSize_{t}$ gradually decreases, which is known to contribute to the convergence of \ac{fl}   \cite{li2019convergence}, 
	it follows from 	Theorem~\ref{thm:DistBound} that the distortion  decreases accordingly, further mitigating its effect as the \ac{fl} iterations progress. 	
	As shown in 
	Appendix \ref{app:Proof1}, the bound in 
	\eqref{eqn:DistBound} is obtained by exploiting the mutual independence of the subtractive dithered quantization error and the quantized value. Hence, our ability to rigorously upper bound the distance in Theorem~\ref{thm:DistBound} is a direct consequence of  this universal  method.

	\vspace{-0.2cm}
	\subsection{\ac{fl} Convergence Analysis}
	\label{subsec:ConvergenceAnalysis}
	\vspace{-0.1cm} 
We next study the convergence of \ac{fl} with \ac{uveqfed}. 
	Our analysis is carried out under the following assumptions, commonly used in  \ac{fl} convergence studies  \cite{stich2018local,li2019convergence}:
	\begin{enumerate}[resume, label={\em AS\arabic*}]
		\item \label{itm:C1} The local objective functions $\{\Objective_k(\cdot)\}$ are all $\SmoothParam$-smooth, namely, for all $\myVec{v}_1, \myVec{v}_2 \in \mySet{R}^{m}$ it holds that
		\begin{equation*}
	\Objective_k(\myVec{v}_1) -  \Objective_k(\myVec{v}_2) \leq (\myVec{v}_1-\myVec{v}_2)^T \nabla \Objective_k(\myVec{v}_2) + \frac{1}{2}\SmoothParam\| \myVec{v}_1-\myVec{v}_2\|^2.
		\end{equation*} 
		\item \label{itm:C2} The local objective functions $\{\Objective_k(\cdot)\}$ are all $\ConvParam$-strongly convex, namely, for all $\myVec{v}_1, \myVec{v}_2 \in \mySet{R}^{m}$ it holds that 
		\begin{equation*}
		\Objective_k(\myVec{v}_1) -  \Objective_k(\myVec{v}_2) \geq (\myVec{v}_1-\myVec{v}_2)^T \nabla \Objective_k(\myVec{v}_2) + \frac{1}{2}\ConvParam\| \myVec{v}_1-\myVec{v}_2\|^2.
		\end{equation*}  
	\end{enumerate}
	Assumptions \ref{itm:C1}-\ref{itm:C2} are commonly used in  \ac{fl} convergence studies  \cite{stich2018local,li2019convergence}, and  hold for a broad range of objective functions used in \ac{fl} systems, including  $\ell_2$-norm regularized linear regression and  logistic regression \cite{li2019convergence}.  
	
	We do not restrict the labeled data of each of the users to be generated from an identical distribution, i.e., 
	we consider a statistically heterogeneous scenario, thus faithfully representing \ac{fl} setups \cite{kairouz2019advances,li2019federated}. 
	Such  heterogeneity is  in line with  assumption \ref{itm:As3}, which does not impose any specific distribution structure on the underlying statistics of the training data. Following \cite{li2019convergence}, we define the heterogeneity gap,
	\begin{equation}
	\label{eqn:HetGap}
	\HetMismatch \triangleq \Objective(\myWeights\Opt) - \sum_{k=1}^{\Nusers}\alpha_{k} \mathop{\min}\limits_{\myWeights}\Objective_k(\myWeights).
	\end{equation}
	The value of $\HetMismatch$ quantifies the degree of heterogeneity. If the training data originates from the same distribution, then  $\HetMismatch$ tends to zero as the training size grows. However, for heterogeneous data, its value is  positive.  The convergence of \ac{uveqfed} with federated averaging is characterized in the following theorem:
	
	\begin{theorem}
		\label{thm:Convergence} 
		Set   $\gamma =\SGDIter\max(1, 4\SmoothParam/  \ConvParam)$ and consider a \ac{uveqfed} setup satisfying \ref{itm:C3}-\ref{itm:C2}.
		Under this setting, local \ac{sgd} with step size $\StepSize_{t} = \frac{ \SGDIter}{\ConvParam\left(t+ \gamma \right) }$  for each $t \in \mySet{N}$   satisfies
		\begin{align}
		&\E\{\Objective(\myWeights_{t})  \} -  \Objective(\myWeights\Opt) \notag \\
		&\qquad \leq \frac{\SmoothParam}{2(t + \gamma)} \max \bigg(\frac{\ConvParam ^2+\SGDIter^2   b}{\SGDIter \ConvParam }, \gamma \|\myWeights_{0} -\myWeights\Opt  \|^2 \bigg),
		\label{eqn:Convergence}
		\end{align}
		where 	
		\begin{equation*}
		 b \triangleq   \left(1 + 4M \ScaleFactor^2 \LatFactor\SGDIter^2 \right)\sum_{k=1}^{\Nusers}\alpha_k^2 \GradVar_k + 6\SmoothParam\HetMismatch + 8(\SGDIter - 1)^2\sum_{k=1}^{\Nusers}\alpha_k \GradMom_k.
		\end{equation*} 
	\end{theorem}
	{\em Proof:}
	See Appendix \ref{app:Proof2}. 
	 
	\smallskip
\textcolor{NewColor}{
Theorem \ref{thm:Convergence} implies that \ac{uveqfed} with local \ac{sgd}, i.e., conventional federated averaging, converges at a rate of $\mySet{O}(1/t)$. The physical meaning of this asymptotic convergence rate is that as the number of iterations $t$ progresses, the learned model converges to the optimal one with a difference decaying as $1/t$. Specifically, the difference between the objective of the model learned in a federated manner and the optimal objective decays to zero at least as quickly as $1/t$ (up to some constant).  This is the same order of convergence as \ac{fl} without quantization constraints for i.i.d. \cite{stich2018local} as well as heterogeneous data \cite{li2019convergence, koloskova2020unified}. Nonetheless, it is noted that the need to quantize the model updates yield an additive term in the coefficient $b$ which grows with the number of parameter via $M$. This term adds to the linear dependence of $b$ on the bound on the gradients norm $\GradMom$, which is expected to grow with the number of parameters, and also appears in the corresponding bounds for local \ac{sgd} without quantization constraints. This implies that \ac{fl} typically converges slower for larger models, i.e.,  the larger the dimensionality of the model updates which have to be quantized. 	A similar order of convergence was also reported for previous probabilistic quantization schemes which typically considered i.i.d. data, e.g., \cite[Thm. 3.4]{alistarh2017qsgd}. 	}
	
	While it is difficult to identify the convergence gains of \ac{uveqfed} over previously proposed \ac{fl} quantizers, such as QSGD, by comparing Theorem \ref{thm:Convergence} to their corresponding convergence bounds, in Section \ref{sec:Sims} we empirically demonstrate that \ac{uveqfed} converges to more accurate global models compared to \ac{fl} with probabilistic scalar quantizers, when trained using i.i.d. as well as heterogeneous data sets. 
	\ifFullVersion
	Additionally, we note that the communication load on the uplink channel induced by \ac{uveqfed} can be further reduced by allowing only part of the nodes to participate in each set of iterations \cite{reisizadeh2019fedpaq,yang2019scheduling}. We leave the analysis of \ac{uveqfed} with partial node participation for future work.  
	\fi
	
	\vspace{-0.2cm}
	\section{Numerical Evaluations}
	\label{sec:Sims}
	\vspace{-0.1cm}
	In this section we numerically evaluate \ac{uveqfed}. We first compare the quantization error induced by \ac{uveqfed} to competing methods utilized in \ac{fl} in Section \ref{subsec:SimsError}. Then, we numerically demonstrate how the reduced distortion is translated in \ac{fl} performance gains using both MNIST and CIFAR-10 data sets\footnote{The source code used in the numerical evaluations detailed in this section is available online at \url{https://github.com/mzchen0/UVeQFed}.} in Section \ref{subsec:SimsConvergence}. 

\vspace{-0.2cm}
\subsection{Quantization Error}
\label{subsec:SimsError}
\vspace{-0.1cm}	
  We begin by focusing only on the compression method, studying its accuracy using synthetic data. We evaluate the distortion induced in quantization of \ac{uveqfed} operating with a two-dimensional hexagonial lattice, i.e., $\Nlattice=2$ and $\GenMat = [2, 0; 1, {1/\sqrt{3}}]$ \cite{kirac1996results}, as well as with scalar quantizers, namely, $\Nlattice=1$ and $\GenMat = 1$. The normalization coefficient is set to $\ScaleFactor = \frac{2+\Rate/5}{\sqrt{M}}$. 
  \textcolor{NewColor}{As discussed in Subsection~\ref{subsec:QuantDiscussion}, decreasing $\ScaleFactor$ results in higher overload probability, i.e., having more quantized sub-vectors lying outside the unit $\Nlattice$-ball. Consequently, the setting used here results in having the quantized sub-vectors being spread in a more uniform manner inside the unit ball at lower quantization rates, where the lattice points are more distant from one another compared to higher quantization rates,  at the cost of increased overload probability, thus balancing the overall distortion.} 
  The distortion of \ac{uveqfed} is compared to QSGD \cite{alistarh2017qsgd}, as well as to uniform quantizers with random unitary rotation \cite{konevcny2016federated}, and to subsampling by random masks followed by uniform three-bit quantizers 
	 \cite{konevcny2016federated}, all operating with the  same quantization rate, i.e., the same    overall number of bits. 

\begin{figure}
	\centering
	\includegraphics[ width=\figWidth, height=\figHeight ]{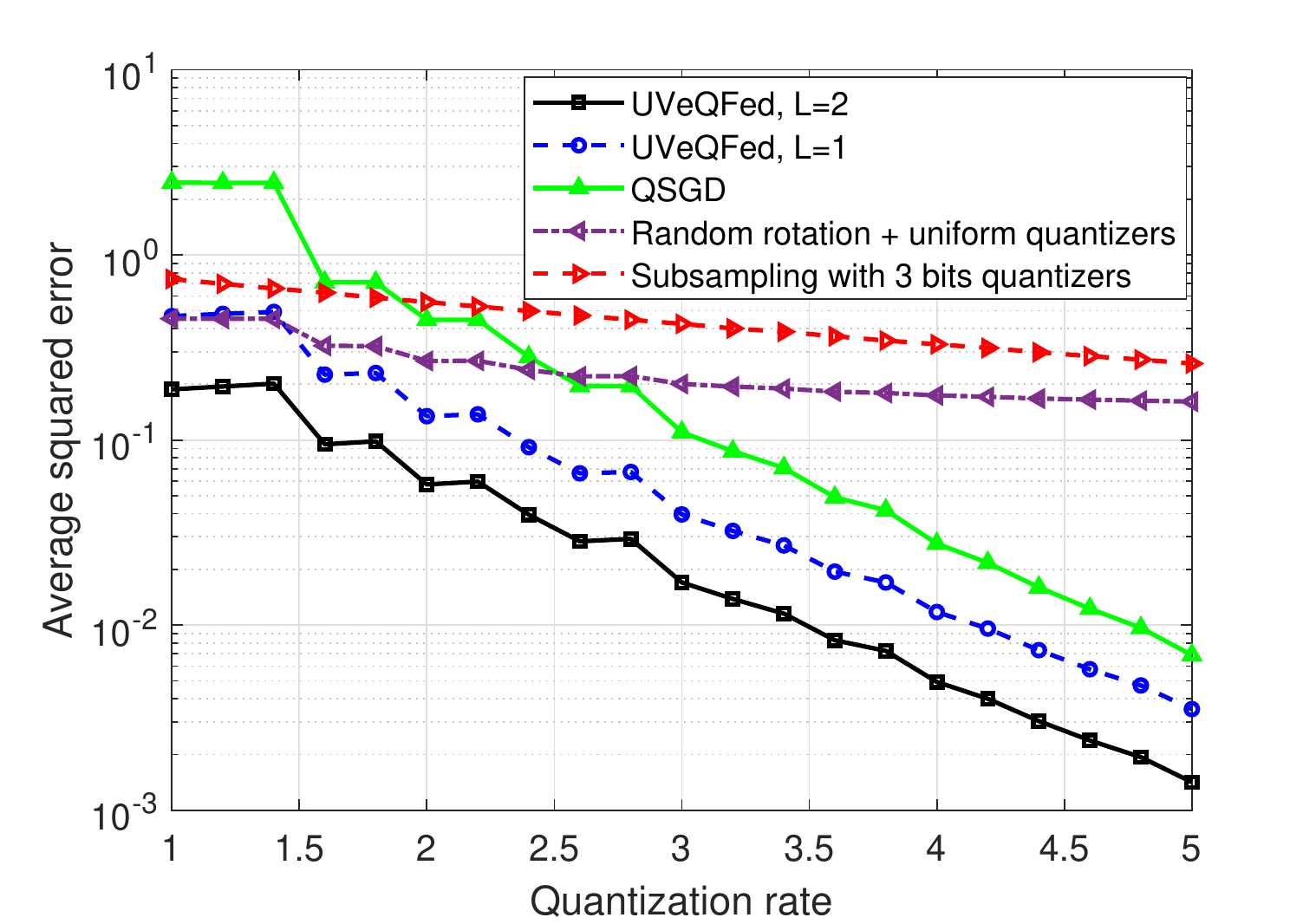}
	\figSpace
	\caption{Quantization distortion, i.i.d. data.
	}
	\label{fig:QuantSchemes1} 
\end{figure}

\begin{figure}
	\centering
	\includegraphics[ width=\figWidth, height=\figHeight]{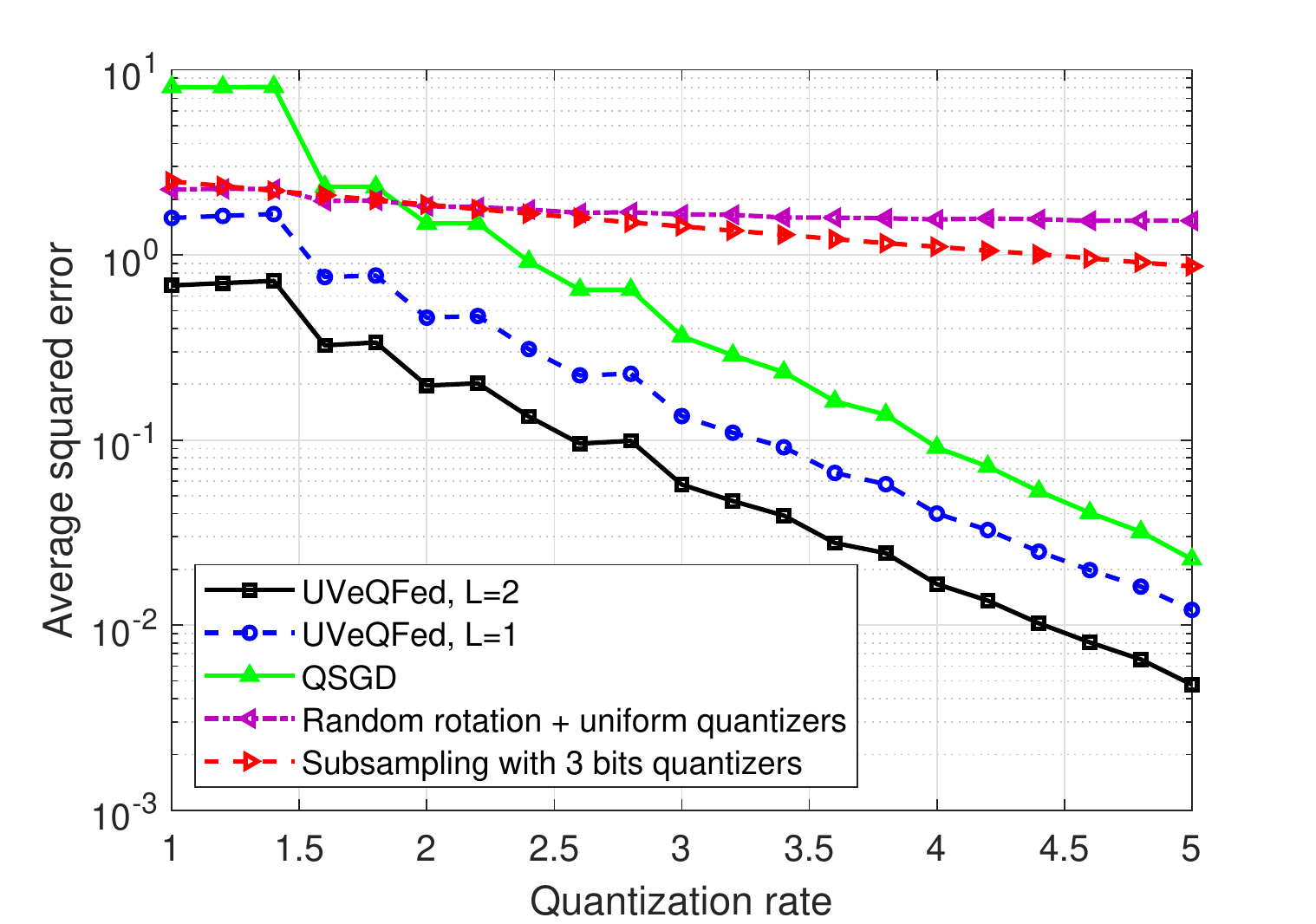}
	\figSpace
	\caption{Quantization distortion, correlated data.
	}
	\label{fig:QuantSchemes2} 
\end{figure}

	Let $\myMat{H}$ be a $128\times 128$ matrix with Gaussian i.i.d. entries, and let $\myMat{\Sigma}$ be a $128\times 128$ matrix whose entries are given by $(\myMat{\Sigma})_{i,j} = e^{-0.2|i-j|}$, representing an exponentially decaying correlation.
	In Figs. \ref{fig:QuantSchemes1}-\ref{fig:QuantSchemes2} we depict the  per-entry squared-error in quantizing $\myMat{H}$ and $\myMat{\Sigma}\myMat{H}\myMat{\Sigma}^T$, representing independent and correlated data, respectively, versus the quantization rate $\Rate$, defined as the ratio of the number of bits to the number of entries of  $\myMat{H}$. The distortion is averaged over $100$ independent realizations of $\myMat{H}$.  To meet the bit rate constraint when using lattice quantizers we scaled $\GenMat$ such that the resulting codewords use less than $128^2 \Rate$ bits. For the scalar quantizers and subsampling-based scheme, the rate determines the quantization resolution and the subsampling ratio, respectively.

	We observe in Figs. \ref{fig:QuantSchemes1}-\ref{fig:QuantSchemes2}  that \ac{uveqfed} achieves a more accurate digital representation compared to  previously proposed methods. It is also observed that \ac{uveqfed} with vector quantization, outperforms its scalar counterpart, and that the gain is more notable when the quantized entries are correlated. This  demonstrates the improved accuracy of jointly encoding multiple samples via vector quantization as well as its ability to exploit statistical correlation in a universal manner by using fixed lattice-based quantization regions which do not depend on the underlying distribution. 

\vspace{-0.2cm}
\subsection{\ac{fl} Convergence}
\label{subsec:SimsConvergence}
\vspace{-0.1cm}	 
  Next, we demonstrate that the reduced distortion of \ac{uveqfed} also translates into \ac{fl} performance gains. To that aim, we evaluate its application for training neural networks  using the MNIST and CIFAR-10 data sets, and compare its performance to that achievable using previous quantization methods for \ac{fl}. 
  \textcolor{NewColor}{The simulation settings are detailed below, with the main parameters summarized in Table~\ref{tbl:Params}.}

\begin{table}
	\color{NewColor}
	\centering
	\caption{Main simulation parameters}
	\label{tbl:Params}
	\begin{tabular}{|p{2cm}|p{1.3cm}| p{1.3cm}|p{2.7cm}|}
		\hline
		&\multicolumn{2}{p{2.6cm}|}{MNIST} & CIFAR-10 \\ \hline
		Users $\Nusers$ & 100 & 15 & 10 \\
	Samples $n_k$ & 500 & 1000 & 5000 \\ \cline{2-3} 
		Model & \multicolumn{2}{p{2.9cm}|}{Two-layer fully connected} & Five-layer convolutional~\cite{matlab2020} \\
		Optimizer &	 \multicolumn{2}{p{2.9cm}|}{Gradient descent}	&	Mini-batch \ac{sgd} \\
		Local steps $\SGDIter$ & \multicolumn{2}{p{2.9cm}|}{ $1$} & $17$ \\
		Step-size $\StepSize_1$ &  \multicolumn{2}{p{2.9cm}|}{$10^{-2}$} & $5\cdot 10^{-3}$\\ \hline
	\end{tabular}
\color{black}	
\end{table}

\begin{figure}
	\centering
	\includegraphics[ width=\figWidth, height=\figHeight]{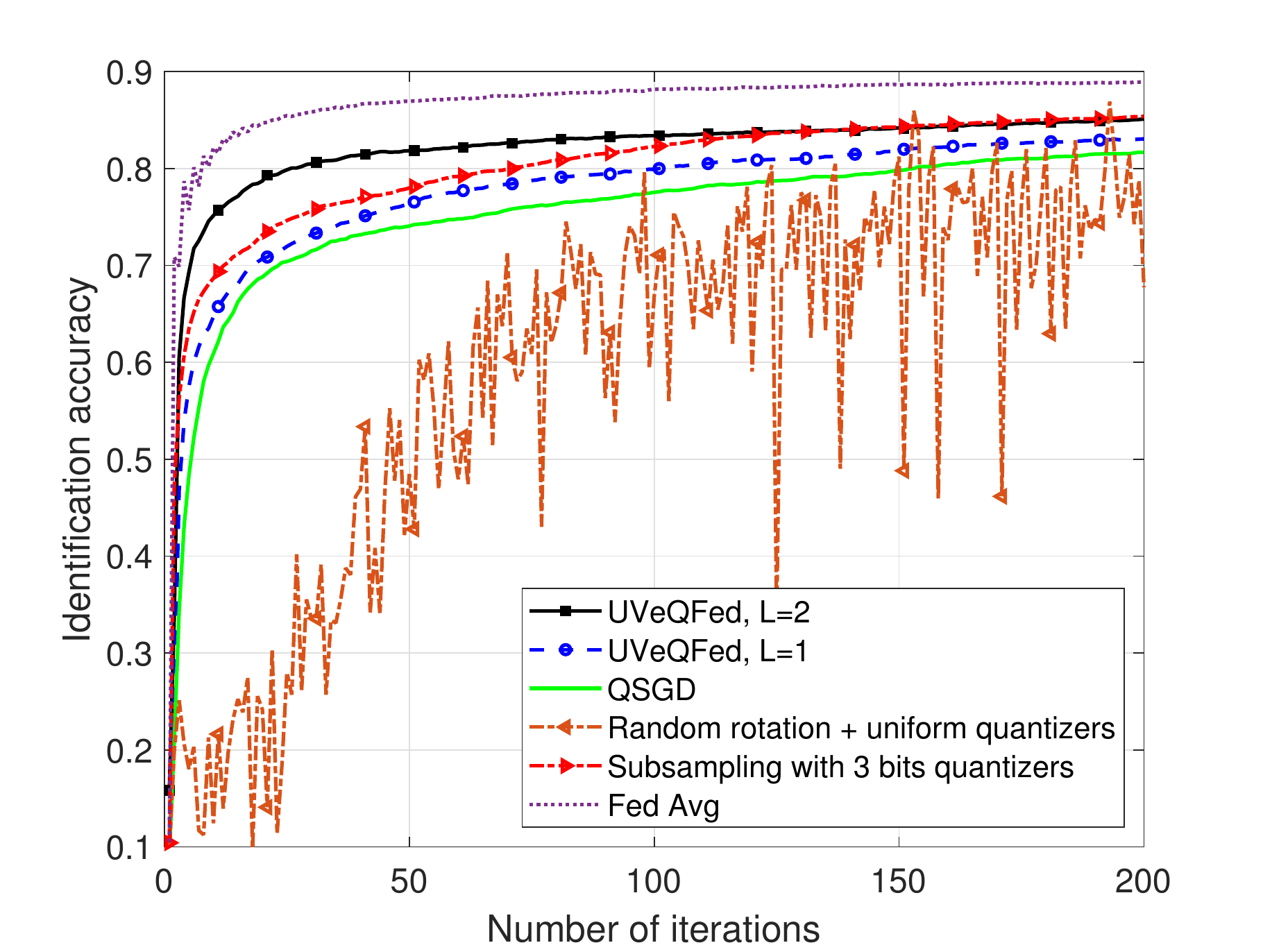}
	\figSpace
	\caption{Convergence profile, MNIST, $\Rate=2$, $\Nusers=100$.
	}
	\label{fig:MNIST_Q2K100} 
\end{figure}

\begin{figure}
	\centering
	\includegraphics[ width=\figWidth, height=\figHeight]{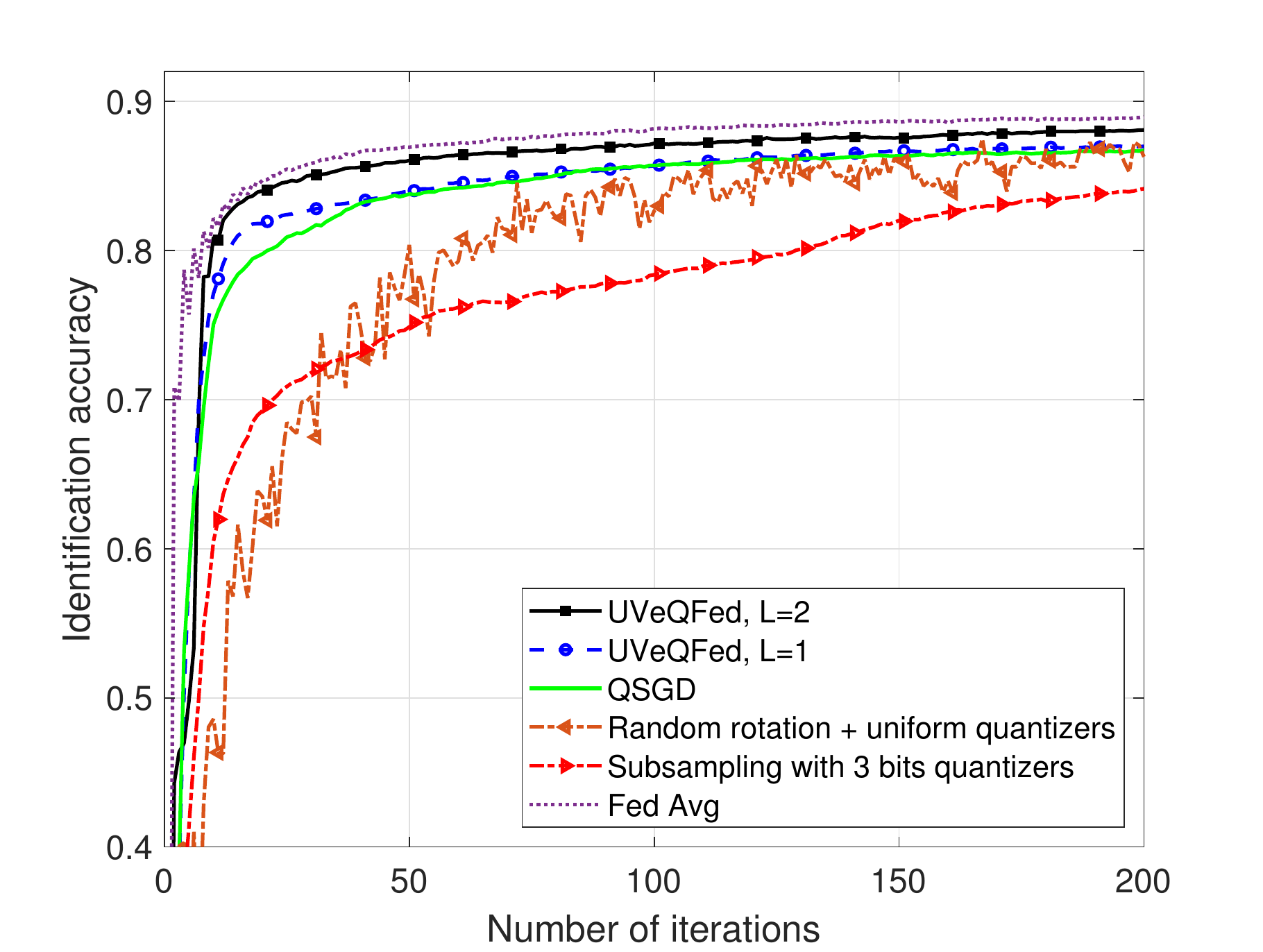}
	\figSpace
	\caption{Convergence profile, MNIST, $\Rate=4$, $\Nusers=100$.
	}
	\label{fig:MNIST_Q4K100} 
\end{figure}

	\color{NewColor}
	We first compare the accuracy of models trained using \ac{uveqfed} to those obtained using federated averaging combined with the quantization methods considered in Subsection~\ref{subsec:SimsError}, i.e., QSGD \cite{alistarh2017qsgd} and the schemes proposed in \cite{konevcny2016federated} of uniform quantizers with random rotation as well as random subsampling followed by three-bit uniform quantizers. 
	To that aim, we train  a fully-connected network with a single hidden layer of $50$ neurons and an intermediate sigmoid activation for detecting handwritten digits based on the MNIST data set. Training is carried out using $\Nusers = 100$ users, each has access to $500$ training samples distributed in an i.i.d. fashion, such that each user has an identical number of images from each label. The users update their weights using gradient descent, where federated averaging is carried out on each iteration.  The resulting accuracy versus the number of iterations of these quantized \ac{fl} schemes compared to federated averaging without quantization is depicted in Figs. \ref{fig:MNIST_Q2K100}-\ref{fig:MNIST_Q4K100} for quantization rates $\Rate = 2$ and $\Rate = 4$, respectively.

Observing Figs.~\ref{fig:MNIST_Q2K100}-\ref{fig:MNIST_Q4K100}, we note that \ac{uveqfed} with vector quantization, i.e., $\Nlattice =2$, achieves the most rapid and accurate convergence among all considered schemes. In particular, for $\Rate = 4$, \ac{uveqfed} with $\Nlattice=2$ achieves a convergence profile within a minor gap from federated averaging without quantization constraints. Among the previous schemes, QSGD demonstrates steady accuracy improvements, though it is still outperformed by \ac{uveqfed} with $\Nlattice=1$, indicating that the reduced distortion achieved by using subtractive dithering is translated into improved trained models. The quantization methods proposed in \cite{konevcny2016federated} result in notable variations in the trained model accuracy and in slower convergence due to their increased error induced in quantization, as noted in Subsection~\ref{subsec:SimsError}.

	We next evaluate \ac{uveqfed} for both heterogeneous as well as i.i.d. distributions of the training data. Based on the results observed in  Figs. \ref{fig:MNIST_Q2K100}-\ref{fig:MNIST_Q4K100} and to avoid cluttering, we compare \ac{uveqfed} only to QSGD and to the accuracy achieved using federated averaging without quantization. Here, we train neural classifiers for both the MNIST and the CIFAR-10 data sets, where for each data set we use both heterogeneous and i.i.d. division of the data.

	\color{black}

\begin{figure}
	\centering
	\includegraphics[ width=\figWidth, height=\figHeight]{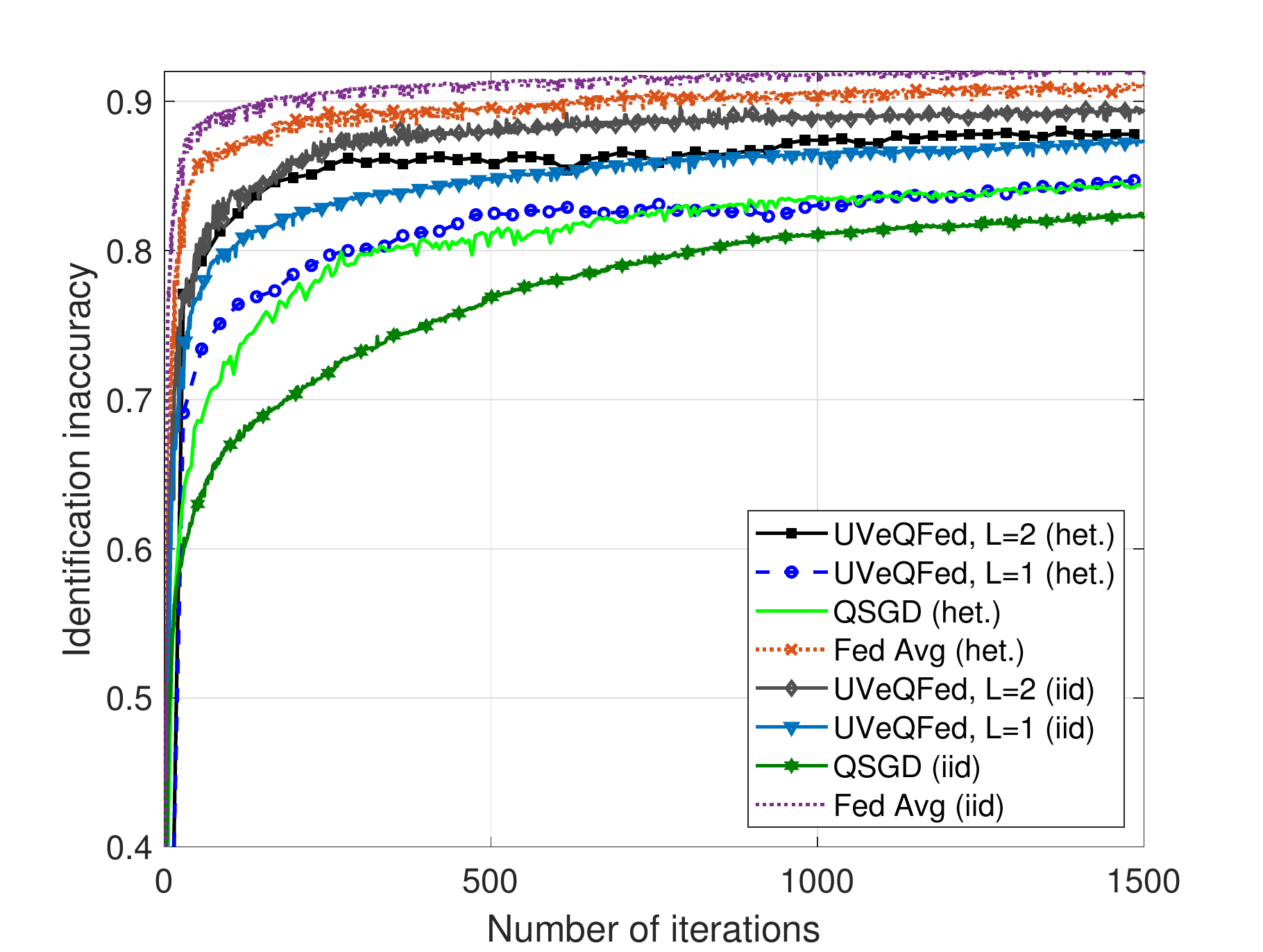}
	\figSpace
	\caption{Convergence profile, MNIST, $\Rate=2$, $\Nusers=15$.
	}
	\label{fig:MNIST_Q2} 
\end{figure}

\begin{figure}
	\centering
	\includegraphics[ width=\figWidth, height=\figHeight]{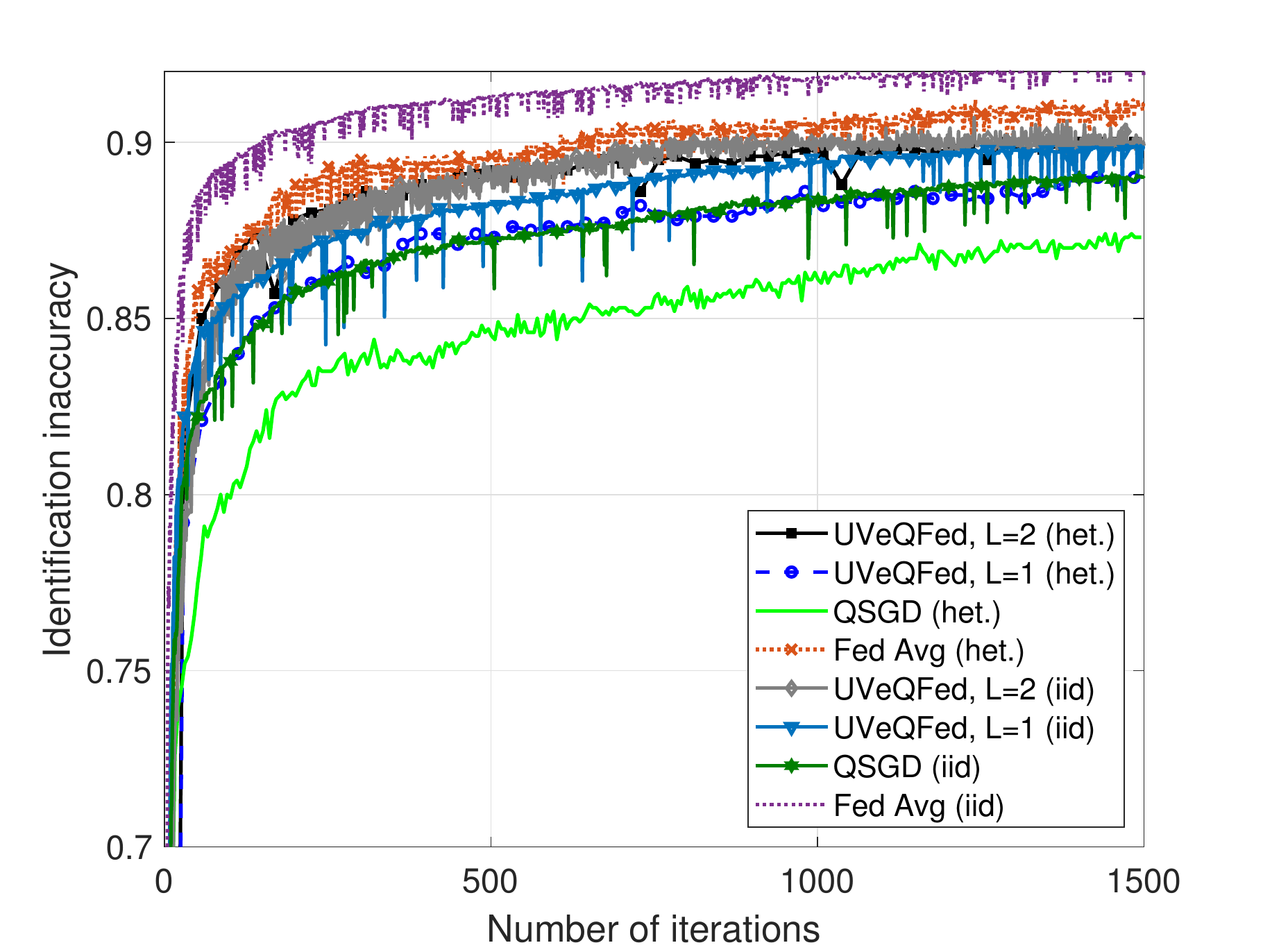}
	\figSpace
	\caption{Convergence profile, MNIST, $\Rate=4$, $\Nusers=15$.
	}
	\label{fig:MNIST_Q4} 
\end{figure}

	For MNIST, we again use a fully-connected network with a single hidden layer of $50$ neurons and an intermediate sigmoid activation with gradient descent optimization. Each of the $\Nusers = 15$ users has  $1000$ training samples. We consider the case where the samples are distributed sequentially among the users, i.e., the first user has the first $1000$ samples in the data set, and so on, resulting in an uneven heterogeneous division of the labels of the users. \textcolor{NewColor}{We also train using an i.i.d. data division, where the labels are uniformly distributed among the users}. 
 The resulting accuracy versus the number of iterations is depicted in Figs. \ref{fig:MNIST_Q2}-\ref{fig:MNIST_Q4} for quantization rates $\Rate = 2$ and $\Rate = 4$, respectively.

\begin{figure}
	\centering
	\includegraphics[ width=\figWidth,height=\figHeight]{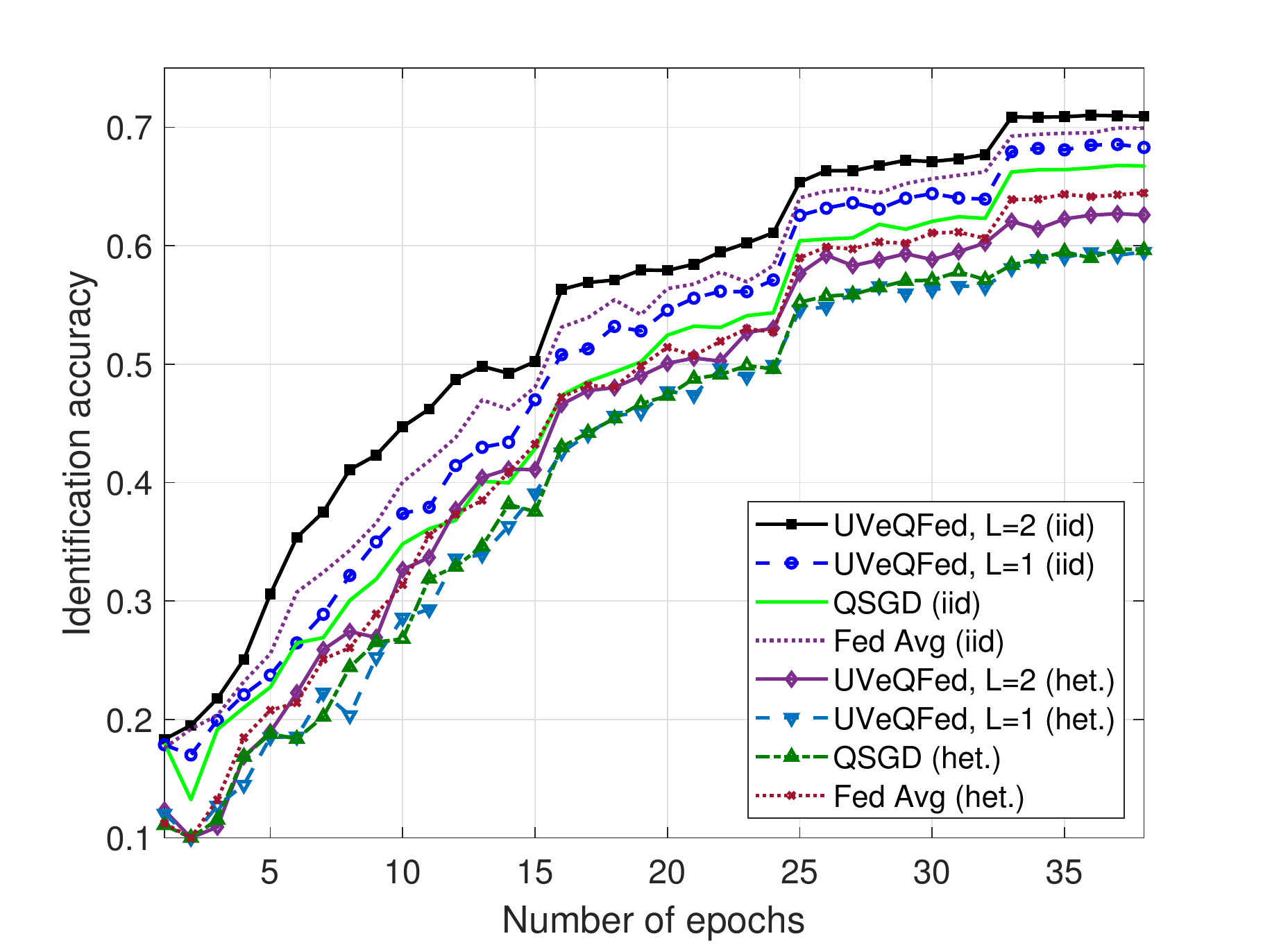}
	\figSpace
	\caption{Convergence profile, CIFAR-10, $\Rate=2$.
	}
	\label{fig:CIFAR_Q2} 
\end{figure}

\begin{figure}
	\centering
	\includegraphics[ width=\figWidth, height=\figHeight]{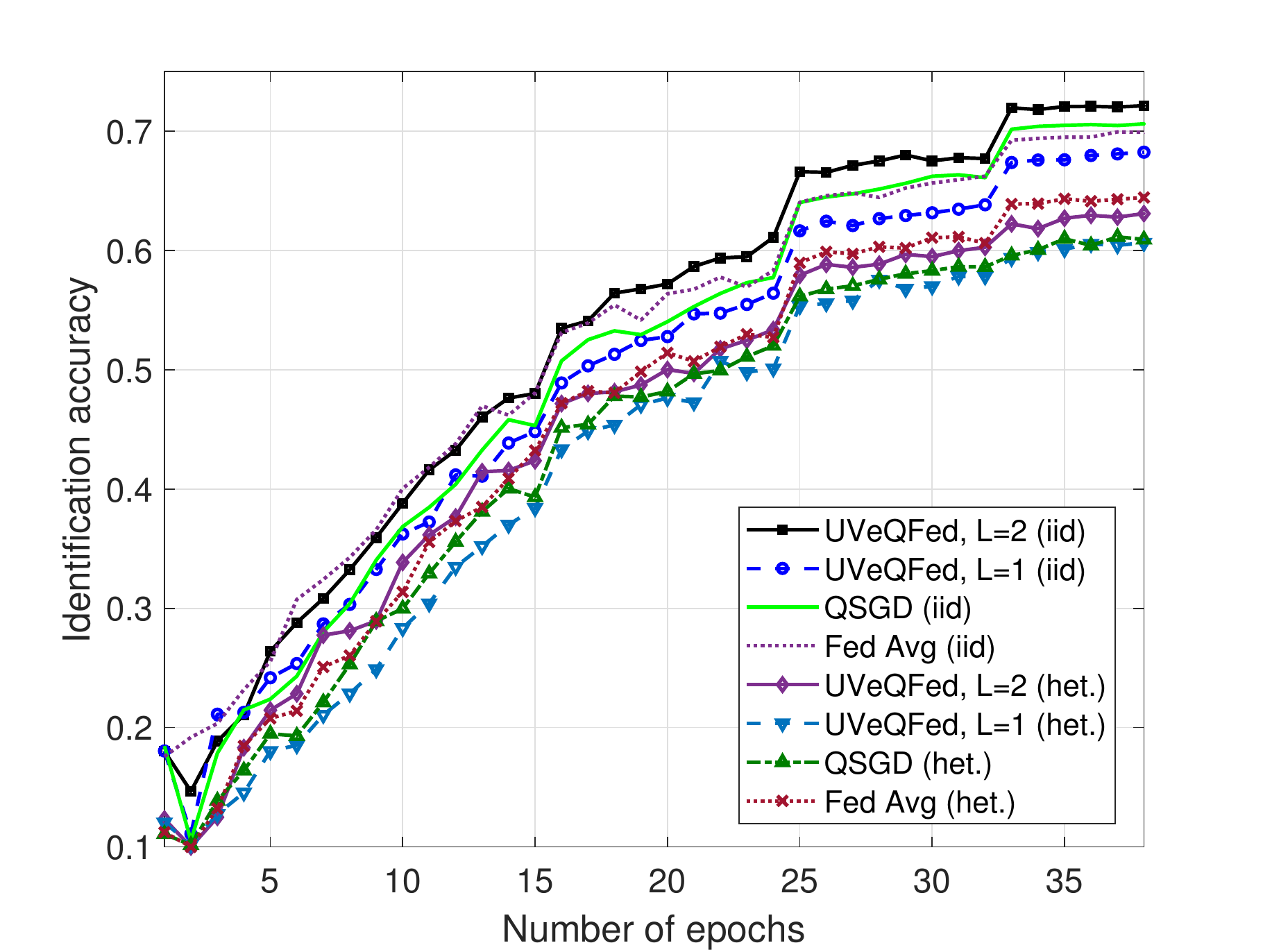}
	\figSpace
	\caption{Convergence profile, CIFAR-10, $\Rate=4$.
	}
	\label{fig:CIFAR_Q4} 
\end{figure}

	For CIFAR-10, we train the deep convolutional neural network architecture used in \cite{matlab2020}, whose trainable parameters constitute  three convolution layers and two fully-connected layers. Here, we consider two methods for distributing the $50000$ training images of CIFAR-10 among the $\Nusers = 10$ users: An i.i.d. division, where each user has the same number  samples from each of the $10$ labels, and a heterogeneous division, in which at least $25\%$ of the  samples of each user correspond to a single distinct label. Each user completes a single epoch of \ac{sgd} with mini-batch size $60$ before the models are aggregated.   The resulting accuracy versus the number of epochs is depicted in Figs. \ref{fig:CIFAR_Q2}-\ref{fig:CIFAR_Q4} for quantization rates $\Rate = 2$ and $\Rate = 4$, respectively.

	We observe in Figs. \ref{fig:MNIST_Q2}-\ref{fig:CIFAR_Q4} that \ac{uveqfed} with vector quantizer, i.e., $\Nlattice = 2$, results in convergence to the most accurate model for all the considered scenarios. \textcolor{NewColor}{In fact, when training a deep convolutional network, for which the loss surface is extremely complex and non-convex, we observe in Figs.~\ref{fig:CIFAR_Q2}-\ref{fig:CIFAR_Q4} that \ac{uveqfed} with $\Nlattice=2$ trained using i.i.d. data achieves improved accuracy over federated averaging without quantization. This follows from the fact that the stochastic nature of the quantization error in \ac{uveqfed} results in its implementing a noisy variant of local \ac{sgd}, which is known to be capable of boosting convergence and avoid local minimas when training \aclp{dnn} with non-convex loss surfaces \cite{Guozhong1995NoiseBackprop}, as also observed in \cite{sery2020over}.} 
	
	   The observed gains are more dominant for $\Rate = 2$, implying that the usage of \ac{uveqfed} with multi-dimensional lattices can notably improve the performance  over low rate channels. 
	Particularly, we observe in Figs. \ref{fig:MNIST_Q2}-\ref{fig:CIFAR_Q4} that similar gains of \ac{uveqfed} are noted for both i.i.d. as well as heterogeneous setups, while the heterogeneous division of the data degrades the accuracy of all considered schemes compared to the i.i.d division. 
	It is also observed that \ac{uveqfed} with scalar quantizers, i.e., $\Nlattice = 1$, achieves improved convergence compared to QSGD for most considered setups, which stems from its reduced distortion. 
	
	The results presented in this section demonstrate that the theoretical benefits of \ac{uveqfed}, which rigorously hold under  \ref{itm:C3}-\ref{itm:C2},  translate into improved convergence when operating under rate constraints with non-synthetic data.

	\vspace{-0.2cm}
	\section{Conclusions}
	\label{sec:Conclusions}
	\vspace{-0.1cm}
	In this work we have proposed \ac{uveqfed}, which utilizes universal vector quantization methods to mitigate the effect of limited communication in \ac{fl}. We first identified the specific requirements from quantization schemes used in \ac{fl} setups. Then, we proposed an encoding-decoding strategy based on dithered lattice quantization. We  analyzed \ac{uveqfed}, proving that its  error term is mitigated by federated averaging. We also characterized its convergence profile, showing that its asymptotic decay rate is the same an unquantized local \ac{sgd}. Our numerical study demonstrates that \ac{uveqfed}  allows achieving more accurate recovery of model updates in each \ac{fl} iteration compared to previously proposed schemes for the same number of bits, and that its reduced distortion is translated into improved convergence with the non-synthetic MNIST and CIFAR-10 data sets.

	\ifFullVersion	
	\begin{appendix}
		\numberwithin{proposition}{subsection} 
		\numberwithin{lemma}{subsection} 
		\numberwithin{corollary}{subsection} 
		\numberwithin{remark}{subsection} 
		\numberwithin{equation}{subsection}

		%
		\vspace{-0.2cm}
		\subsection{Proof of Theorem \ref{thm:QError}}
		\label{app:Proof0}
		To prove the theorem, we note that by decoding step \ref{itm:Collect}, the error vector $\Qerror_{t+\SGDIter}\IdxK$ scaled by $\ScaleFactor\|\myUpdate_{t+\SGDIter}\IdxK\|$, consists of $M$ vectors $\{\QerrorV_i\IdxK\}$. Each $\QerrorV_i\IdxK$ is an $\Nlattice \times 1$ vector representing the $i$th subtractive dithered quantization error, defined as $\QerrorV_i\IdxK \triangleq   Q_\mySet{L}(\myUpdateV_i\IdxK +\myVec{z}_i\IdxK  )- \myVec{z}_i\IdxK - \myUpdateV_i\IdxK$. 	The fact that we have used subtractive dithered quantization via encoding steps \ref{itm:Dither}-\ref{itm:Quantize} and decoding step \ref{itm:SubDither}, implies that, regardless of the statistical model of $\{\myUpdateV_i\IdxK \}$, the  quantization error vectors  $\{\QerrorV_i\IdxK\}$ are zero-mean, i.i.d (over both $i$ and $k$), and uniformly distributed over $\Partition$ \cite{zamir1996lattice}. Consequently, 
		\begin{align*}
		\E\left\{\left\|\Qerror_{t+\SGDIter}\IdxK \right\|^2 \big| \myUpdate_{t+\SGDIter}\IdxK \right\} 
		&= \ScaleFactor^2\|\myUpdate_{t+\SGDIter}\IdxK\|^2\sum_{i=1}^{M}\E\left\{\left\|\QerrorV_i\IdxK \right\|^2   \right\} \notag \\
		&= \ScaleFactor^2\|\myUpdate_{t+\SGDIter}\IdxK\|^2 M \LatFactor,
		\end{align*}
		thus proving the theorem. 
		\qed
		
		\vspace{-0.2cm}
		\subsection{Proof of Theorem \ref{thm:DistBound}}
		\label{app:Proof1}
		To prove the the theorem, we first express the the updated global model using as a sum of the desired global model and the quantization noise. Then, we show that the distance between $\myWeights_{t+\SGDIter}$ and    $\myWeights_{t+\SGDIter}\Desired$ can be bounded via \eqref{eqn:DistBound} due to the statistical properties of subtractive dithered quantization error \cite{zamir1996lattice}.  
		To formulate this distance between $\myWeights_{t+\SGDIter}$ and the desired  $\myWeights_{t+\SGDIter}\Desired$,  we use
		$\{\myWeightsV_{t,i}\}_{i=1}^{\Npoints}$,  $\{\myGlobalModelV_{t+\SGDIter, i}\}_{i=1}^{\Npoints}$, and $\{\myGlobalModelV_{t+\SGDIter, i}\Desired\}_{i=1}^{\Npoints}$ to denote the partitions of $\myWeights_{t}$, $\myWeights_{t+\SGDIter}$,  and $\myWeights_{t+\SGDIter}\Desired$ into $\Npoints$ distinct $\Nlattice \times 1$ vectors, as done in Step \ref{itm:Partition}.
		To formulate this distance,  we use 
		$\{\myWeightsV_{t,i}\}_{i=1}^{\Npoints}$ to denote the  partition of $\myWeights_t$ into $\Npoints$ distinct $\Nlattice \times 1$ vectors via step \ref{itm:Partition}, similarly to the definitions of $		\{\myGlobalModelV_{t+\SGDIter, i}\}$ and $\{\myGlobalModelV_{t+\SGDIter, i}\Desired\}$.
		
		From the decoding and model recovery steps \ref{itm:Collect}-\ref{itm:Recovery}  it follows that 
		\vspace{-0.1cm}
		\begin{align}
		&\myGlobalModelV_{t+\SGDIter, i} 
		= \myWeightsV_{t,i} + \sum\limits_{k=1}^{\Nusers} \alpha_k \ScaleFactor\|\myUpdate_{t+\SGDIter}\IdxK\|  \left(  Q_\mySet{L}(\myUpdateV_i\IdxK +\myVec{z}_i\IdxK  )- \myVec{z}_i\IdxK \right) \notag  \\
		&=  \myWeightsV_{t,i} + \sum\limits_{k=1}^{\Nusers} \alpha_k \ScaleFactor\|\myUpdate_{t+\SGDIter}\IdxK\| \myUpdateV_i\IdxK + \sum\limits_{k=1}^{\Nusers} \alpha_k \ScaleFactor\|\myUpdate_{t+\SGDIter}\IdxK\| \QerrorV_i\IdxK, 
		\vspace{-0.1cm}
		\label{eqn:Proof1Eq1}
		\end{align}
		where $\QerrorV_i\IdxK$ is the subtractive dithered quantization error, defined in Appendix \ref{app:Proof0}. Now, since $\myUpdate_{t+\SGDIter}\IdxK = \myGlobalModel_{t+\SGDIter}\IdxK - \myWeights_t$ combined with \eqref{eqn:UpdatedModel} and the fact that $\sum_{k=1}^{\Nusers} \alpha_k = 1$, it holds that $\myWeightsV_{t,i} + \sum_{k=1}^{\Nusers} \alpha_k \ScaleFactor\|\myUpdate_{t+\SGDIter}\IdxK\|\myUpdateV_i\IdxK = \myGlobalModelV_{t+\SGDIter, i}\Desired$. Substituting this into  \eqref{eqn:Proof1Eq1} yields
		\vspace{-0.1cm}
		\begin{equation}
		\myGlobalModelV_{t+\SGDIter, i} - \myGlobalModelV_{t+\SGDIter, i}\Desired = \sum\limits_{k=1}^{\Nusers} \alpha_k \ScaleFactor\|\myUpdate_{t+\SGDIter}\IdxK\|\QerrorV_i\IdxK. 
		\label{eqn:Dist1}
		\vspace{-0.1cm}
		\end{equation}
		
		As discussed in Appendix \ref{app:Proof0}, $\{\QerrorV_i\IdxK\}$ are zero-mean, i.i.d (over both $i$ and $k$), and independent of $\myUpdate_{t+\SGDIter}\IdxK$. Consequently, by the law of total expectation
			\begin{align}
	&\E\left\{ \left\|\myWeights_{t+\SGDIter} - \myWeights_{t+\SGDIter}\Desired\right\|^2   \right\} 
	= \E\left\{ \left\| \sum\limits_{i=1}^{M} \sum\limits_{k=1}^{\Nusers} \alpha_k \ScaleFactor\|\myUpdate_{t+\SGDIter}\IdxK\|\QerrorV_i\IdxK\right\|^2   \right\} \notag \\
	&\stackrel{(a)}{=} \E\left\{ \E\left\{\left\| \sum\limits_{i=1}^{M} \sum\limits_{k=1}^{\Nusers} \alpha_k \ScaleFactor\|\myUpdate_{t+\SGDIter}\IdxK\|\QerrorV_i\IdxK\right\|^2 \Big| \myUpdate_{t+\SGDIter}\IdxK \right\}  \right\} \notag \\
	&\stackrel{(b)}{=} \E\left\{ M\sum\limits_{k=1}^{\Nusers} \alpha_k^2 \ScaleFactor^2 \LatFactor \|\myUpdate_{t+\SGDIter}\IdxK\|^2\right\}. 
	\label{eqn:Dist1a}
	\vspace{-0.1cm}
	\end{align}		
	where $(a)$ follows from the law of total expectation, and $(b)$ holds by \eqref{eqn:QError}.

	Next, we note that by \eqref{eqn:SGD1}, the model update $\myUpdate_{t+\SGDIter}\IdxK = \myGlobalModel_{t+\SGDIter}\IdxK - 	\myGlobalModel_{t}\IdxK$ can be written as  the sum of the stochastic gradients $\myUpdate_{t+\SGDIter}\IdxK  = \sum_{t'=t}^{t+\SGDIter-1}\StepSize_{t'} \nabla\Objective_k^{i_{t'}\IdxK}\big(\myGlobalModel_{t'}\IdxK  \big)$. Since the indices $\{i_{t}\IdxK\}$ are i.i.d. over $t$ and $k$, applying the law of total expectation  to \eqref{eqn:Dist1a}  yields
	\begin{align}
	&\E\left\{ \left\|\myWeights_{t+\SGDIter} - \myWeights_{t+\SGDIter}\Desired\right\|^2   \right\}\notag \\ 
&=\!\E\left\{\! M\ScaleFactor^2 \LatFactor\sum\limits_{k=1}^{\Nusers} \alpha_k^2  \E\left\{\|\myUpdate_{t+\SGDIter}\IdxK\|^2 \Big|\{\myGlobalModel_{t'}\IdxK \} \right\} \right\} \notag \\
&=\! \E\left\{\! M\ScaleFactor^2 \LatFactor\sum\limits_{k=1}^{\Nusers}\! \alpha_k^2  \E\left\{ \left\|\sum_{t'=t}^{t+\SGDIter - 1}\!\!\StepSize_{t'} \nabla\Objective_k^{i_{t'}\IdxK}\!\!\big(\myGlobalModel_{t'}\IdxK  \big)  \right\|^2\!\Big|\!\{\myGlobalModel_{t'}\IdxK \}\! \right\} \!\right\}  \notag \\
		&\stackrel{(a)}{\leq} \! \E\left\{\! M\ScaleFactor^2 \LatFactor\sum\limits_{k=1}^{\Nusers}\! \alpha_k^2  \SGDIter\!\! \sum_{t'=t}^{t+\SGDIter - 1}\!\! \StepSize_{t'}^2	\E\left\{ \left\|\nabla\Objective_k^{i_{t'}\IdxK}\!\!\big(\myGlobalModel_{t'}\IdxK  \big)  \right\|^2 \! \Big|\!\{\myGlobalModel_{t'}\IdxK \} \!\right\} \!\right\} \notag \\
&\stackrel{(b)}{\leq} \!
M\ScaleFactor^2 \LatFactor \SGDIter \left( \sum_{t'=t}^{t+\SGDIter - 1} \StepSize_{t'}^2\right)  \sum\limits_{k=1}^{\Nusers} \alpha_k^2\GradMom_k , 
\label{eqn:Dist1b}
\end{align}	
where in $(a)$ we used the inequality $\|\sum_{t'={t+1-\SGDIter}}^{t+1} \myVec{r}_t\|^2 \leq \SGDIter\sum_{t'=t+1-\SGDIter}^{t+1}  \|\myVec{r}_t\|^2$, which holds for any multivariate sequence $\{\myVec{r}_t\}$; and $(b)$ holds since the uniform distribution of the random index $i_k$ implies that the expected value of the stochastic gradient is the full gradient, i.e., $\E\big\{\nabla\Objective_k^{i_t\IdxK}\big(\myWeights  \big) \big\} = \nabla\Objective_k(\myWeights)$, and consequently, $\E\big\{\big\|\nabla\Objective_k^{i_t\IdxK}\big(\myGlobalModel_{t}\IdxK  \big)\! -\! \nabla\Objective_k\big(\myGlobalModel_{t}\IdxK \big)  \big\|^2  \big\} \leq \E\big\{\big\|\nabla\Objective_k^{i_t\IdxK}\big(\myGlobalModel_{t}\IdxK \big)   \big\|^2  \big\} \leq \GradMom_k$ by \ref{itm:C3}. 		
Equation \eqref{eqn:Dist1b}  proves the theorem.
		\qed

		\vspace{-0.2cm}
		\subsection{Proof of Theorem \ref{thm:Convergence}}
		\label{app:Proof2}
		\vspace{-0.1cm}
		Our proof follows a similar outline to that used in \cite{stich2018local,li2019convergence}, with the introduction of additional arguments for handling  the quantization constraints. The unique characteristics of the quantization error which arise from the dithered strategy presented in Section \ref{sec:Quantization} allow us to rigorously incorporate its contribution into the overall flow of the proof.

		\subsubsection{Recursive Bound on Weights Error}
		\label{app:Proof2a}
		
		From \cite{zamir1996lattice} it follows that the effect of substractive dithered quantization can be modeled as additive noise, independent of the quantized value, whose distribution depends only on the properties of the lattice.  In particular, it holds that 
		 the distortion induced in quantizing the model update $\myUpdate_{t}\IdxK$, denoted $\Qerror_{t}\IdxK$,  is an $m \times 1$ zero-mean additive noise vector {\em independent of $\myUpdate_{t\SGDIter}\IdxK$, and thus also of $\myGlobalModel_{t}\IdxK$ and $i_t\IdxK$}. 
		  Consequently, by defining the sequence $\QerrorSeq_t\IdxK$ such that $\QerrorSeq_t\IdxK =\Qerror_{t}\IdxK $ if $t$ is an integer multiple of $\SGDIter$ and $\QerrorSeq_t\IdxK =\myVec{0} $ otherwise, it follows that  \eqref{eqn:SGD1} can be written as
\ifsingle
\begin{align}
\label{eqn:SGD2}
&\myGlobalModel_{t+1}\IdxK = 	\begin{cases}
\myGlobalModel_{t}\IdxK\! -\! \StepSize_t \nabla\Objective_k^{i_t\IdxK}\big(\myGlobalModel_{t}\IdxK ) \big)\! +\! \QerrorSeq_{t+1}\IdxK  & t\!+\!1  \!\notin\!  \mySet{T}_{\SGDIter}, \\
\sum\limits_{k'=1}^{\Nusers}\!\alpha_{k'}\!  \left(\! \myGlobalModel_{t}\IdxKt \!\!-\! \StepSize_t \nabla\Objective_k^{i_t\IdxKt}\!\big(\myGlobalModel_{t}\IdxKt  \big) \!+\! \QerrorSeq_{t+1}\IdxKt \right) & t\!+\!1  \!\in\! \mySet{T}_{\SGDIter}.
\end{cases}
\end{align}
\else
		\begin{align}
		\label{eqn:SGD2}
		&\myGlobalModel_{t+1}\IdxK = \\ 
	&	\begin{cases}
		\myGlobalModel_{t}\IdxK\! -\! \StepSize_t \nabla\Objective_k^{i_t\IdxK}\big(\myGlobalModel_{t}\IdxK ) \big)\! +\! \QerrorSeq_{t+1}\IdxK  & t\!+\!1  \!\notin\!  \mySet{T}_{\SGDIter}, \\
		\sum\limits_{k'=1}^{\Nusers}\!\alpha_{k'}\!  \left(\! \myGlobalModel_{t}\IdxKt \!\!-\! \StepSize_t \nabla\Objective_k^{i_t\IdxKt}\!\big(\myGlobalModel_{t}\IdxKt  \big) \!+\! \QerrorSeq_{t+1}\IdxKt \right) & t\!+\!1  \!\in\! \mySet{T}_{\SGDIter}.
		\end{cases}\notag 
		\end{align}
\fi
		
		The equivalent model update representation \eqref{eqn:SGD2} allows us to model the effect of subtractive dithered quantization on the overall \ac{fl} procedure as additional noise corrupting the computation of the stochastic gradients. Building upon this representation, we now follow the strategy proposed in \cite{stich2018local} and adapted to heterogeneous data in \cite{li2019convergence}. This is achieved by defining a virtual sequence $\{\myVec{v}_t\}$ from $\{\myGlobalModel_{t}\IdxK\}$ which can be shown to behave almost like mini-batch \ac{sgd} with batch size $\SGDIter$, while being within a bounded distance of the \ac{fl} model weights  $\{\myGlobalModel_{t}\IdxK\}$, by properly setting the step size $\StepSize_t$.  In particular, we define the virtual sequence $\{\myVec{v}_t\}$ via 
		\begin{equation}
		\label{eqn:VirSGD}
		\myVec{v}_t \triangleq \sum_{k=1}^{\Nusers}\alpha_k \myGlobalModel_{t}\IdxK,
		\end{equation}
		which coincides with $\myGlobalModel_{t}\IdxK$ when $t$ is an integer multiple of $\SGDIter$. Further define the averaged noisy stochastic gradients and the averaged full gradients as
		\begin{subequations} 
		\label{eqn:GradDef}
		\begin{align}
		\label{eqn:GradRand}
		\GradRand_t &\triangleq \sum_{k=1}^{\Nusers}\alpha_k\left(  \nabla\Objective_k^{i_t\IdxK}\big(\myGlobalModel_{t}\IdxK \big) -\frac{1}{\StepSize_t}\QerrorSeq_{t+1}\IdxK \right) , \\
		 \GradDet_t &\triangleq \sum_{k=1}^{\Nusers}\alpha_k \nabla\Objective_k\big(\myGlobalModel_{t}\IdxK \big), 
		 \label{eqn:GradDet}
		\end{align}
	\end{subequations}
		respectively. Note that since the quantization error is zero-mean and the sample indexes $\{i_t\IdxK\}$ are independent and uniformly distributed, it holds that $\E\{\GradRand_t\} = \GradDet_t$. Additionally, the virtual sequence \eqref{eqn:VirSGD} satisfies	$\myVec{v}_{t+1} = 	\myVec{v}_t -\StepSize_t \GradRand_t$. 
		
		The resulting model is thus equivalent to that used in \cite[App. A]{li2019convergence}, and as a result, by assumptions \ref{itm:C1}-\ref{itm:C2}, it follows from \cite[Lemma 1]{li2019convergence} that if $\StepSize_t \leq \frac{1}{4\SmoothParam}$ then 
\ifsingle	
		\begin{align}
		\E\left\{\left\| \myVec{v}_{t+1} \!-\! \myWeights\Opt\right\|^2 \right\} 
		&\leq (1\!-\!\StepSize_t\ConvParam) \E\left\{\left\| \myVec{v}_{t} \!-\! \myWeights\Opt\right\|^2 \right\} \!  + \!6 \SmoothParam \StepSize_t^2 \HetMismatch \notag \\
		& \!	+ \!\StepSize_t^2 \E\left\{\left\|	\GradRand_t\! - \!	\GradDet_t \right\|^2  \right\}  \!+ \!2\E\left\{ \sum_{k=1}^{\Nusers}\!\alpha_k\! \left\|\myVec{v}_{t} \!- \!	\myGlobalModel_{t}\IdxK \right\|^2  \right\}.
		\label{eqn:Proof2Stp1}  
		\end{align}
\else		
		\begin{align}
		&\E\left\{\left\| \myVec{v}_{t+1} \!-\! \myWeights\Opt\right\|^2 \right\} 
		\leq (1\!-\!\StepSize_t\ConvParam) \E\left\{\left\| \myVec{v}_{t} \!-\! \myWeights\Opt\right\|^2 \right\} \!  + \!6 \SmoothParam \StepSize_t^2 \HetMismatch \notag \\
		& \!	+ \!\StepSize_t^2 \E\left\{\left\|	\GradRand_t\! - \!	\GradDet_t \right\|^2  \right\}  \!+ \!2\E\left\{ \sum_{k=1}^{\Nusers}\!\alpha_k\! \left\|\myVec{v}_{t} \!- \!	\myGlobalModel_{t}\IdxK \right\|^2  \right\}.
		\label{eqn:Proof2Stp1}  
		\end{align}
\fi
		Expression \eqref{eqn:Proof2Stp1} bounds the expected distance between the virtual sequence $\{\myVec{v}_t\}$ and the optimal weights $\myWeights\Opt$ in a recursive manner. We further bound the summands in \eqref{eqn:Proof2Stp1}, using the following lemmas: 
		\begin{lemma}
			\label{lem:GradBound} 
			If the step size $\StepSize_t$ is non-increasing and satisfies $\StepSize_t \leq 2 \StepSize_{t+\SGDIter}$ for each $t\geq 0$, then, when assumption \ref{itm:C3} is satisfied, it holds that  
			\begin{equation}
			\StepSize_t^2 \E\left\{\left\|	\GradRand_t - 	\GradDet_t \right\|^2  \right\}  \leq \left(1 + 4M \ScaleFactor^2\LatFactor\SGDIter^2 \right)\StepSize_t^2 \sum_{k=1}^{\Nusers}\alpha_k^2 \GradVar_k .
			\label{eqn:GradBound}
			\end{equation}	
		\end{lemma}
		%
		
		\begin{lemma}
			\label{lem:SumBound}
			If the step size $\StepSize_t$ is non-increasing and satisfies $\StepSize_t \leq 2 \StepSize_{t+\SGDIter}$ for each $t\geq 0$, then, by \ref{itm:C3}, it holds that
			\begin{equation}
			\label{eqn:SumBound}
			\E\left\{ \sum_{k=1}^{\Nusers}\alpha_k \left\|\myVec{v}_{t} - 	\myGlobalModel_{t}\IdxK \right\|^2  \right\} \leq 4(\SGDIter - 1)^2\StepSize_{t}^2\sum_{k=1}^{\Nusers}\alpha_k \GradMom_k.
			\end{equation}
		\end{lemma}

		Next, we define  $\delta_t \triangleq  \E\left\{\left\| \myVec{v}_{t} - \myWeights\Opt\right\|^2 \right\} $. When $t \in \mySet{T}_{\SGDIter}$, the term $\delta_t $ represents the $\ell_2$ norm of the error in the weights of the global model. Using Lemmas \ref{lem:GradBound}-\ref{lem:SumBound}, while substituting \eqref{eqn:SumBound} and \eqref{eqn:GradBound} into \eqref{eqn:Proof2Stp1},   we obtain  the  following recursive relationship on the weights error:
		\begin{equation}
		\delta_{t+1} \leq (1- \StepSize_{t}\ConvParam) \delta_t + \StepSize_{t}^2 b,
		\label{eqn:RecursiveRel}
		\end{equation}
		where 
		\begin{equation*}
		b \triangleq   \left(1 + 4M \ScaleFactor^2\LatFactor\SGDIter^2 \right)\sum_{k=1}^{\Nusers}\alpha_k^2 \GradVar_k + 6\SmoothParam\HetMismatch + 8(\SGDIter - 1)^2\sum_{k=1}^{\Nusers}\alpha_k \GradMom_k. 
		\end{equation*}

		The relationship in \eqref{eqn:RecursiveRel} is used in the sequel to prove the \ac{fl} convergence bound stated in Theorem \ref{thm:Convergence}.
		
		\subsubsection{\ac{fl} Convergence Bound}
		\label{app:Proof2b}
		Here, we prove Theorem \ref{thm:Convergence} based on the recursive relationship in \eqref{eqn:RecursiveRel}. This is achieved by properly setting the step-size and the \ac{fl} systems parameters in \eqref{eqn:RecursiveRel} to bound $\delta_{t} = \E\left\{\left\| \myVec{v}_{t} - \myWeights\Opt\right\|^2 \right\} $, and combining the resulting bound with the strong convexity of the objective \ref{itm:C2} to prove \eqref{eqn:Convergence}. 
		
		In particular, we set the step size $\StepSize_{t}$ to take the form $\StepSize_{t} = \frac{\beta}{t + \gamma}$ for some $\beta > 0$ and $\gamma \geq \max\big(4 \SmoothParam \beta, \SGDIter\big)$, for which $\StepSize_{t} \leq \frac{1}{4\SmoothParam}$ and $\StepSize_{t} \leq 2\StepSize_{t+\SGDIter}$, implying that \eqref{eqn:Proof2Stp1} and \eqref{eqn:SumBound} hold.
		
		Under such settings, we show that there exists a finite $\nu$ such that $\delta_{t}  \leq \frac{\nu}{t + \gamma}$ for all integer $l \geq 0$. We prove this by induction, noting that setting $\nu \geq\gamma \delta_{0}$ guarantees that it holds for $t=0$. Consequently, we next show that if  $\delta_{t}  \leq \frac{\nu}{t + \gamma}$, then  $\delta_{t+1}  \leq \frac{\nu}{t+1 + \gamma}$. It follows from  \eqref{eqn:RecursiveRel} that 
		\begin{align}
		\delta_{t+1} &\leq \left(1- \frac{\beta}{ t + \gamma}\ConvParam\right)  \frac{\nu}{t + \gamma}+ \left( \frac{\beta}{ t+ \gamma}\right)^2  b  \notag \\
		&=   \frac{1}{t+\SGDIter}\left(\left(1- \frac{\beta}{ t + \gamma}\ConvParam\right)\nu + \frac{\beta^2}{ t+ \gamma}  b \right) . 
		\label{eqn:ConvProof1}
		\end{align}
		Consequently,  $\delta_{t+1}  \leq \frac{\nu}{t+1 + \gamma}$ holds when  
		\begin{equation*}
		\frac{1}{t+\SGDIter}\left(\left(1- \frac{\beta}{ t + \gamma}\ConvParam\right)\nu + \frac{\beta^2}{ t+ \gamma}  b \right) 
		\leq \frac{\nu}{t+1 + \gamma},
		\end{equation*}
		or, equivalently, 
		\begin{equation}
		\left(1- \frac{\beta}{ t + \gamma}\ConvParam\right)\nu + \frac{\beta^2}{ t+ \gamma}  b \leq \frac{t+ \gamma}{t+1 + \gamma}\nu.
		\label{eqn:ConvProof3}
		\end{equation}
		By setting   ${\nu} \geq \frac{1+\beta^2   b}{\beta \ConvParam }$, the left hand side of \eqref{eqn:ConvProof3} satisfies
\ifsingle	
	\begin{align}
	\left(1\!-\! \frac{\beta}{ t\! + \!\gamma}\ConvParam\right)\nu \!+\! \frac{\beta^2}{ t\!+\! \gamma} b
	&\!=\! \frac{ t\!-\!1\!+\! \gamma}{t\!+\! \gamma}{\nu} \!+ \! \left(\frac{1\!-\!\beta\ConvParam}{t\!+\!\gamma}\nu\! +\!\frac{\beta^2 }{t\!+\!\gamma}b \right)  \notag \\
	&= \frac{ t-1+ \gamma}{t+ \gamma}{\nu} \!+ \! \frac{1}{{t+\gamma}} \left(\left( {1-\beta\ConvParam}\right) \nu +\beta^2 b \right) \notag \\
	&\stackrel{(a)}{\leq}\frac{ t-1+ \gamma}{t+ \gamma}{\nu},
	\label{eqn:ConvProof4}
	\end{align}
\else
		\begin{align}
		&\left(1\!-\! \frac{\beta}{ t\! + \!\gamma}\ConvParam\right)\nu \!+\! \frac{\beta^2}{ t\!+\! \gamma} b
		\!=\! \frac{ t\!-\!1\!+\! \gamma}{t\!+\! \gamma}{\nu} \!+ \! \left(\frac{1\!-\!\beta\ConvParam}{t\!+\!\gamma}\nu\! +\!\frac{\beta^2 }{t\!+\!\gamma}b \right)  \notag \\
		&= \frac{ t-1+ \gamma}{t+ \gamma}{\nu} \!+ \! \frac{1}{{t+\gamma}} \left(\left( {1-\beta\ConvParam}\right) \nu +\beta^2 b \right) \notag \\
		&\stackrel{(a)}{\leq}\frac{ t-1+ \gamma}{t+ \gamma}{\nu},
		\label{eqn:ConvProof4}
		\end{align}
\fi
		where $(a)$ holds since ${\nu} \geq \frac{1+\beta^2  b}{\beta \ConvParam }$. As the right hand side of \eqref{eqn:ConvProof4} is not larger than that of \eqref{eqn:ConvProof3}, it follows that \eqref{eqn:ConvProof3} holds for the current setting, which in turn proves that  $\delta_{t+1}  \leq \frac{\nu}{t+1 + \gamma}$. 
		Finally, the smoothness of the objective \ref{itm:C1} implies that
		\begin{equation}
		\E\{\Objective(\myWeights_{t})  \} -  \Objective(\myWeights\Opt) \leq \frac{\SmoothParam}{2}\delta_{t}\leq \frac{\SmoothParam \nu}{2(t+ \gamma)},
		\label{eqn:ConvProof5}
		\end{equation}
		which, in light of the above setting, holds for $\nu \ge \max \big(\frac{1+\beta^2   b}{\beta \ConvParam }, \gamma \delta_{0} \big)$, $\gamma \ge \max(\SGDIter, 4\beta \SmoothParam)$, and $\beta > 0$. In particular, setting $\beta =   \frac{\SGDIter}{\ConvParam}$ results in  $\gamma \ge  \SGDIter\max(1, 4\SmoothParam/  \ConvParam)$ and  $\nu \ge \max \big(\frac{\ConvParam ^2+\SGDIter^2   b}{\SGDIter \ConvParam }, \gamma \delta_{0} \big)$, which, when substituted into \eqref{eqn:ConvProof5}, proves \eqref{eqn:Convergence}. 		
			\qed

		\subsubsection{Deferred Proofs}
		\label{app:Proof2c}
		Here we detail the proofs of the intermediate lemmas used for obtaining the recursion \eqref{eqn:RecursiveRel}.
		\paragraph{Proof of Lemma \ref{lem:GradBound}}
		To prove \eqref{eqn:GradBound}, we note that since the quantization noise and the stochastic gradients are mutually independent, it follows from the definition of the gradient vectors \eqref{eqn:GradDef} that
\ifsingle	
\begin{align}
\StepSize_t^2 \E\left\{\left\|	\GradRand_t - 	\GradDet_t \right\|^2  \right\}   &= \sum_{k=1}^{\Nusers}\alpha_k^2  \E\left\{\left\| \QerrorSeq_{t+1}\IdxK \right\|^2  \right\}  + \StepSize_t^2 \sum_{k=1}^{\Nusers}\alpha_k^2 \E\left\{\left\|\nabla\Objective_k^{i_t\IdxK}\big(\myGlobalModel_{t}\IdxK  \big)\! -\! \nabla\Objective_k\big(\myGlobalModel_{t}\IdxK \big)  \right\|^2  \right\}  
\notag \\ 
& \stackrel{(a)}{\leq} \sum_{k=1}^{\Nusers}\alpha_k^2  \E\left\{\left\| \QerrorSeq_{t+1}\IdxK \right\|^2  \right\} + \StepSize_t^2 \sum_{k=1}^{\Nusers}\alpha_k^2 \GradVar_k ,
\label{eqn:GradBound1}
\end{align}
\else		
		\begin{align}
		&\StepSize_t^2 \E\left\{\left\|	\GradRand_t - 	\GradDet_t \right\|^2  \right\}   = \sum_{k=1}^{\Nusers}\alpha_k^2  \E\left\{\left\| \QerrorSeq_{t+1}\IdxK \right\|^2  \right\} \notag \\
		& \qquad\qquad + \StepSize_t^2 \sum_{k=1}^{\Nusers}\alpha_k^2 \E\left\{\left\|\nabla\Objective_k^{i_t\IdxK}\big(\myGlobalModel_{t}\IdxK  \big)\! -\! \nabla\Objective_k\big(\myGlobalModel_{t}\IdxK \big)  \right\|^2  \right\}  
		 \notag \\ 
		& \qquad\stackrel{(a)}{\leq} \sum_{k=1}^{\Nusers}\alpha_k^2  \E\left\{\left\| \QerrorSeq_{t+1}\IdxK \right\|^2  \right\} + \StepSize_t^2 \sum_{k=1}^{\Nusers}\alpha_k^2 \GradVar_k ,
		\label{eqn:GradBound1}
		\end{align}
\fi				
		where $(a)$   holds since the uniform distribution of the random index $i_k$ implies that the expected value of the stochastic gradient is the full gradient, i.e., $\E\big\{\nabla\Objective_k^{i_t\IdxK}\big(\myWeights  \big) \big\} = \nabla\Objective_k(\myWeights)$, and consequently, $\E\big\{\big\|\nabla\Objective_k^{i_t\IdxK}\big(\myGlobalModel_{t}\IdxK  \big)\! -\! \nabla\Objective_k\big(\myGlobalModel_{t}\IdxK \big)  \big\|^2  \big\} \leq \E\big\{\big\|\nabla\Objective_k^{i_t\IdxK}\big(\myGlobalModel_{t}\IdxK \big)   \big\|^2  \big\} \leq \GradMom_k$ by \ref{itm:C3}. 				
		Furthermore,  the definition of $\QerrorSeq_{t+1}\IdxK$ implies that $\E\big\{\big\| \QerrorSeq_{t+1}\IdxK \big\|^2  \big\} = 0$ for $t+1 \notin \mySet{T}$, while for  $t+1 \in \mySet{T}$ it holds that $\E\big\{\big\| \QerrorSeq_{t+1}\IdxK \big\|^2  \big\} = \E\big\{\big\| \Qerror_{t+1}\IdxK \big\|^2  \big\}  = M \sigma_{\mySet{L}_{t+1}\IdxK}^2$. 
		Now, similarly to the derivation in \eqref{eqn:Dist1b},  the quantization error induced by \ac{uveqfed} satisfies
		\begin{align}
		&\E\big\{\big\| \QerrorSeq_{t+1}\IdxK \big\|^2  \big\} 
		\leq M \ScaleFactor^2 \LatFactor \E\left\{ \left\|\sum_{t'=t+1-\SGDIter}^{t+1} \StepSize_{t'}\nabla\Objective_k^{i_{t'}\IdxK}\big(\myGlobalModel_{t'}\IdxK  \big)  \right\|^2  \right\} \notag \\
		&\stackrel{(a)}{\leq} M \ScaleFactor^2\LatFactor \SGDIter \sum_{t'=t+1-\SGDIter}^{t+1} \StepSize_{t'}^2	\E\left\{ \left\|\nabla\Objective_k^{i_{t'}\IdxK}\big(\myGlobalModel_{t'}\IdxK  \big)  \right\|^2  \right\} \notag \\
		&\stackrel{(b)}{\leq} M\ScaleFactor^2 \LatFactor \SGDIter^2\StepSize_{t+1-\SGDIter}^2 \GradMom_k 
		\stackrel{(c)}{\leq} 4M \ScaleFactor^2\LatFactor \SGDIter^2\StepSize_{t}^2 \GradMom_k, 
		\label{eqn:GradBound1a}
		\end{align}	
			where in $(a)$ we used the inequality $\|\sum_t'={t+1-\SGDIter}^{t+1} \myVec{r}_t\|^2 \leq \SGDIter\sum_{t'=t+1-\SGDIter}^{t+1}  \|\myVec{r}_t\|^2$, which holds for any multivariate sequence $\{\myVec{r}_t\}$; $(b)$ is obtained from assumption \ref{itm:C3}; and $(c)$ follows since $ \StepSize_{t+1 - \SGDIter} \leq 2 \StepSize_{t+1} \leq 2 \StepSize_{t}$. 
			  Substituting \eqref{eqn:GradBound1a} into \eqref{eqn:GradBound1} proves the lemma.		
		\qed
		
		\paragraph{Proof of Lemma \ref{lem:SumBound}}
		Note that for $t_0 = \lfloor t/ \SGDIter \rfloor\SGDIter$, which is an integer multiple of $\SGDIter$, it holds that $\myVec{v}_{t_0} = 	\myGlobalModel_{t_0}\IdxK$. Since  \eqref{eqn:SumBound} trivially holds for $t=t_0$, we henceforth focus on the case where $t > t_0$. We now write 
		\begin{align}
		&\E\left\{ \sum_{k=1}^{\Nusers}\alpha_k \left\|\myGlobalModel_{t}\IdxK \! - \! \myVec{v}_{t}  \right\|^2  \right\} \notag \\
		&\qquad = 		\E\left\{ \sum_{k=1}^{\Nusers}\alpha_k \left\|\myGlobalModel_{t}\IdxK \! - \! 	\myGlobalModel_{t_0}\IdxK \! - \! (\myVec{v}_{t} \! - \! \myVec{v}_{t_0})  \right\|^2  \right\} \notag \\
		&\qquad \stackrel{(a)}{\leq} 	\E\left\{ \sum_{k=1}^{\Nusers}\alpha_k \left\|\myGlobalModel_{t}\IdxK \! - \! 	\myGlobalModel_{t_0}\IdxK \right\|^2  \right\} \notag \\
		&\qquad =  \sum_{k=1}^{\Nusers}\alpha_k  	\E\left\{\left\|\myGlobalModel_{t}\IdxK \! - \! 	\myGlobalModel_{t_0}\IdxK \right\|^2  \right\},
		\label{eqn:ProoflemAid1}
		\end{align}
		where in $(a)$ we used the fact that for every set $\{\myVec{r}\IdxK\}$, one can define a random vector $\myVec{r}$ such that $\Pr(\myVec{r}= \myVec{r}\IdxK) = \alpha_k$, and thus 
		\begin{align*}
		\sum_{k=1}^{\Nusers} \alpha_{k} \Big\|\myVec{r}\IdxK - \sum_{l=1}^{\Nusers} \alpha_l \myVec{r}^{(l)}\Big\|^2 &= \E\{\|\myVec{r} - \E\{\myVec{r}\}\|^2 \} \notag \\
		&\leq \E\{\|\myVec{r}\|^2 \} =	\sum_{k=1}^{\Nusers} \alpha_k \|\myVec{r}\IdxK\|^2.
		\end{align*}
		
		Next, we recall that $\QerrorSeq_{t'} = \myVec{0}$ for each $t' = t_0 + 1, \ldots, t$.  Consequently, similarly to the derivation in \eqref{eqn:GradBound1a},  
\ifsingle
\begin{align}
\E\left\{ \left\|\myGlobalModel_{t}\IdxK - 	\myGlobalModel_{t_0}\IdxK \right\|^2  \right\} &= 	\E\left\{ \left\|\sum_{t'=t_0}^{t-1} \StepSize_{t'}\nabla\Objective_k^{i_{t'}\IdxK}\big(\myGlobalModel_{t'}\IdxK  \big)  \right\|^2  \right\} \notag \\
&\stackrel{(a)}{\leq} (\SGDIter - 1)\sum_{t'=t_0}^{t-1} \StepSize_{t'}^2	\E\left\{ \left\|\nabla\Objective_k^{i_{t'}\IdxK}\big(\myGlobalModel_{t'}\IdxK  \big)  \right\|^2  \right\} \notag \\
&\stackrel{(b)}{\leq}(\SGDIter - 1)^2\StepSize_{t_0}^2 \GradMom_k
\stackrel{(c)}{\leq}4(\SGDIter - 1)^2\StepSize_{t}^2 \GradMom_k,
\label{eqn:ProoflemAid2}
\end{align}
\else
		\begin{align}
		&\E\left\{ \left\|\myGlobalModel_{t}\IdxK - 	\myGlobalModel_{t_0}\IdxK \right\|^2  \right\} = 	\E\left\{ \left\|\sum_{t'=t_0}^{t-1} \StepSize_{t'}\nabla\Objective_k^{i_{t'}\IdxK}\big(\myGlobalModel_{t'}\IdxK  \big)  \right\|^2  \right\} \notag \\
		&\stackrel{(a)}{\leq} (\SGDIter - 1)\sum_{t'=t_0}^{t-1} \StepSize_{t'}^2	\E\left\{ \left\|\nabla\Objective_k^{i_{t'}\IdxK}\big(\myGlobalModel_{t'}\IdxK  \big)  \right\|^2  \right\} \notag \\
		&\stackrel{(b)}{\leq}(\SGDIter - 1)^2\StepSize_{t_0}^2 \GradMom_k
		\stackrel{(c)}{\leq}4(\SGDIter - 1)^2\StepSize_{t}^2 \GradMom_k,
		\label{eqn:ProoflemAid2}
		\end{align}
\fi		
		where in $(a)$ we used the inequality $\|\sum_{t'=t_0}^{t-1} \myVec{r}_t\|^2 \leq (t-1-t_0)\sum_{t'=t_0}^{t-1}  \|\myVec{r}_t\|^2 \leq (\SGDIter-1)\sum_{t'=t_0}^{t-1}  \|\myVec{r}_t\|^2$, which holds for any multivariate sequence $\{\myVec{r}_t\}$; $(b)$ is obtained from assumption \ref{itm:C3}; and $(c)$ follows since $\StepSize_{t_0} \leq \StepSize_{t - \SGDIter} \leq 2 \StepSize_t$.  Substituting \eqref{eqn:ProoflemAid2} into \eqref{eqn:ProoflemAid1} proves the lemma.	
		\qed


	\end{appendix}
\fi
\bibliographystyle{IEEEtran}
\bibliography{IEEEabrv,refs}

\begin{thebibliography}{10}
\providecommand{\url}[1]{#1}
\csname url@samestyle\endcsname
\providecommand{\newblock}{\relax}
\providecommand{\bibinfo}[2]{#2}
\providecommand{\BIBentrySTDinterwordspacing}{\spaceskip=0pt\relax}
\providecommand{\BIBentryALTinterwordstretchfactor}{4}
\providecommand{\BIBentryALTinterwordspacing}{\spaceskip=\fontdimen2\font plus
\BIBentryALTinterwordstretchfactor\fontdimen3\font minus
  \fontdimen4\font\relax}
\providecommand{\BIBforeignlanguage}[2]{{%
\expandafter\ifx\csname l@#1\endcsname\relax
\typeout{** WARNING: IEEEtran.bst: No hyphenation pattern has been}%
\typeout{** loaded for the language `#1'. Using the pattern for}%
\typeout{** the default language instead.}%
\else
\language=\csname l@#1\endcsname
\fi
#2}}
\providecommand{\BIBdecl}{\relax}
\BIBdecl

\bibitem{shlezinger2020federated}
N.~Shlezinger, M.~Chen, Y.~C. Eldar, H.~V. Poor, and S.~Cui, ``Federated
  learning with quantization constraints,'' in \emph{Proc. IEEE Int. Conf.
  Acoust. Speech Signal Process.}, 2020, pp. 8851--8855.

\bibitem{lecun2015deep}
Y.~LeCun, Y.~Bengio, and G.~Hinton, ``Deep learning,'' \emph{Nature}, vol. 521,
  no. 7553, p. 436, 2015.

\bibitem{chen2019deep}
J.~Chen and X.~Ran, ``Deep learning with edge computing: A review,''
  \emph{Proceedings of the IEEE}, 2019.

\bibitem{dean2012large}
J.~Dean, G.~Corrado, R.~Monga, K.~Chen, M.~Devin, M.~Mao, A.~Senior, P.~Tucker,
  K.~Yang, Q.~V. Le \emph{et~al.}, ``Large scale distributed deep networks,''
  in \emph{Neural Information Processing Systems}, 2012, pp. 1223--1231.

\bibitem{mcmahan2016communication}
H.~B. McMahan, E.~Moore, D.~Ramage, and S.~Hampson, ``Communication-efficient
  learning of deep networks from decentralized data,'' \emph{arXiv preprint
  arXiv:1602.05629}, 2016.

\bibitem{bonawitz2019towards}
K.~Bonawitz, H.~Eichner, W.~Grieskamp, D.~Huba, A.~Ingerman, V.~Ivanov,
  C.~Kiddon, J.~Kone{\v{c}}n{\`y}, S.~Mazzocchi, and H.~B. McMahan, ``Towards
  federated learning at scale: System design,'' \emph{arXiv preprint
  arXiv:1902.01046}, 2019.

\bibitem{kairouz2019advances}
P.~Kairouz, H.~B. McMahan, B.~Avent, A.~Bellet, M.~Bennis, A.~N. Bhagoji,
  K.~Bonawitz, Z.~Charles, G.~Cormode, R.~Cummings \emph{et~al.}, ``Advances
  and open problems in federated learning,'' \emph{arXiv preprint
  arXiv:1912.04977}, 2019.

\bibitem{chen2019joint}
M.~Chen, Z.~Yang, W.~Saad, C.~Yin, H.~V. Poor, and S.~Cui, ``A joint learning
  and communications framework for federated learning over wireless networks,''
  \emph{arXiv preprint arXiv:1909.07972}, 2019.

\bibitem{li2019federated}
T.~Li, A.~K. Sahu, A.~Talwalkar, and V.~Smith, ``Federated learning:
  Challenges, methods, and future directions,'' \emph{arXiv preprint
  arXiv:1908.07873}, 2019.

\bibitem{yang2019scheduling}
H.~H. Yang, Z.~Liu, T.~Q. Quek, and H.~V. Poor, ``Scheduling policies for
  federated learning in wireless networks,'' \emph{arXiv preprint
  arXiv:1908.06287}, 2019.

\bibitem{amiri2020update}
M.~M. Amiri, D.~Gunduz, S.~R. Kulkarni, and H.~V. Poor, ``Update aware device
  scheduling for federated learning at the wireless edge,'' \emph{arXiv
  preprint arXiv:2001.10402}, 2020.

\bibitem{konevcny2016federated}
J.~Kone{\v{c}}n{\`y}, H.~B. McMahan, F.~X. Yu, P.~Richt{\'a}rik, A.~T. Suresh,
  and D.~Bacon, ``Federated learning: Strategies for improving communication
  efficiency,'' \emph{arXiv preprint arXiv:1610.05492}, 2016.

\bibitem{lin2017deep}
Y.~Lin, S.~Han, H.~Mao, Y.~Wang, and W.~J. Dally, ``Deep gradient compression:
  Reducing the communication bandwidth for distributed training,'' \emph{arXiv
  preprint arXiv:1712.01887}, 2017.

\bibitem{hardy2017distributed}
C.~Hardy, E.~Le~Merrer, and B.~Sericola, ``Distributed deep learning on
  edge-devices: feasibility via adaptive compression,'' in \emph{Proc.
  International Symposium on Network Computing and Applications (NCA)}.\hskip
  1em plus 0.5em minus 0.4em\relax IEEE, 2017, pp. 1--8.

\bibitem{aji2017sparse}
A.~F. Aji and K.~Heafield, ``Sparse communication for distributed gradient
  descent,'' \emph{arXiv preprint arXiv:1704.05021}, 2017.

\bibitem{wen2017terngrad}
W.~Wen, C.~Xu, F.~Yan, C.~Wu, Y.~Wang, Y.~Chen, and H.~Li, ``Terngrad: Ternary
  gradients to reduce communication in distributed deep learning,'' in
  \emph{Neural Information Processing Systems}, 2017, pp. 1509--1519.

\bibitem{alistarh2017qsgd}
D.~Alistarh, D.~Grubic, J.~Li, R.~Tomioka, and M.~Vojnovic, ``{QSGD}:
  Communication-efficient {SGD} via gradient quantization and encoding,'' in
  \emph{Neural Information Processing Systems}, 2017, pp. 1709--1720.

\bibitem{horvath2019natural}
S.~Horvath, C.-Y. Ho, L.~Horvath, A.~N. Sahu, M.~Canini, and P.~Richtarik,
  ``Natural compression for distributed deep learning,'' \emph{arXiv preprint
  arXiv:1905.10988}, 2019.

\bibitem{reisizadeh2019fedpaq}
A.~Reisizadeh, A.~Mokhtari, H.~Hassani, A.~Jadbabaie, and R.~Pedarsani,
  ``Fedpaq: A communication-efficient federated learning method with periodic
  averaging and quantization,'' in \emph{International Conference on Artificial
  Intelligence and Statistics}, 2020, pp. 2021--2031.

\bibitem{horvath2019stochastic}
S.~Horv{\'a}th, D.~Kovalev, K.~Mishchenko, S.~Stich, and P.~Richt{\'a}rik,
  ``Stochastic distributed learning with gradient quantization and variance
  reduction,'' \emph{arXiv preprint arXiv:1904.05115}, 2019.

\bibitem{bernstein2018signsgd}
J.~Bernstein, Y.-X. Wang, K.~Azizzadenesheli, and A.~Anandkumar, ``{SignSGD}:
  Compressed optimisation for non-convex problems,'' \emph{arXiv preprint
  arXiv:1802.04434}, 2018.

\bibitem{polyanskiy2014lecture}
Y.~Polyanskiy and Y.~Wu, ``Lecture notes on information theory,'' \emph{Lecture
  Notes for 6.441 (MIT), ECE563 (University of Illinois Urbana-Champaign), and
  STAT 664 (Yale)}, 2012-2017.

\bibitem{zamir1992universal}
R.~Zamir and M.~Feder, ``On universal quantization by randomized
  uniform/lattice quantizers,'' \emph{{IEEE} Trans. Inf. Theory}, vol.~38,
  no.~2, pp. 428--436, 1992.

\bibitem{zamir1996lattice}
------, ``On lattice quantization noise,'' \emph{{IEEE} Trans. Inf. Theory},
  vol.~42, no.~4, pp. 1152--1159, 1996.

\bibitem{li2019convergence}
X.~Li, K.~Huang, W.~Yang, S.~Wang, and Z.~Zhang, ``On the convergence of fedavg
  on non-iid data,'' \emph{arXiv preprint arXiv:1907.02189}, 2019.

\bibitem{speedtest2019}
\BIBentryALTinterwordspacing
speedtest.net, ``Speedtest united states market report,'' 2019. [Online].
  Available: \url{http://www.speedtest.net/reports/united-states/}
\BIBentrySTDinterwordspacing

\bibitem{gray1998quantization}
R.~M. Gray and D.~L. Neuhoff, ``Quantization,'' \emph{{IEEE} Trans. Inf.
  Theory}, vol.~44, no.~6, pp. 2325--2383, 1998.

\bibitem{chou1996vector}
P.~A. Chou, M.~Effros, and R.~M. Gray, ``A vector quantization approach to
  universal noiseless coding and quantization,'' \emph{{IEEE} Trans. Inf.
  Theory}, vol.~42, no.~4, pp. 1109--1138, 1996.

\bibitem{ziv1985universal}
J.~Ziv, ``On universal quantization,'' \emph{{IEEE} Trans. Inf. Theory},
  vol.~31, no.~3, pp. 344--347, 1985.

\bibitem{gray1993dithered}
R.~M. Gray and T.~G. Stockham, ``Dithered quantizers,'' \emph{{IEEE} Trans.
  Inf. Theory}, vol.~39, no.~3, pp. 805--812, 1993.

\bibitem{lipshitz1992quantization}
S.~P. Lipshitz, R.~A. Wannamaker, and J.~Vanderkooy, ``Quantization and dither:
  A theoretical survey,'' \emph{Journal of the Audio Engineering Society},
  vol.~40, no.~5, pp. 355--375, 1992.

\bibitem{zamir1999multiterminal}
R.~Zamir and T.~Berger, ``Multiterminal source coding with high resolution,''
  \emph{{IEEE} Trans. Inf. Theory}, vol.~45, no.~1, pp. 106--117, 1999.

\bibitem{kirac1996results}
A.~Kirac and P.~Vaidyanathan, ``Results on lattice vector quantization with
  dithering,'' \emph{{IEEE} Trans. Circuits Syst. {II}}, vol.~43, no.~12, pp.
  811--826, 1996.

\bibitem{conway2013sphere}
J.~H. Conway and N.~J.~A. Sloane, \emph{Sphere Packings, Lattices and
  Groups}.\hskip 1em plus 0.5em minus 0.4em\relax Springer Science \& Business
  Media, 2013, vol. 290.

\bibitem{rubinstein1982generating}
R.~Rubinstein, ``Generating random vectors uniformly distributed inside and on
  the surface of different regions,'' \emph{European Journal of Operational
  Research}, vol.~10, no.~2, pp. 205--209, 1982.

\bibitem{aysal2008distributed}
T.~C. Aysal, M.~J. Coates, and M.~G. Rabbat, ``Distributed average consensus
  with dithered quantization,'' \emph{{IEEE} Trans. Signal Process.}, vol.~56,
  no.~10, pp. 4905--4918, 2008.

\bibitem{shlezinger2020task}
N.~Shlezinger and Y.~C. Eldar, ``Task-based quantization with application to
  {MIMO} receivers,'' \emph{arXiv preprint arXiv:2002.04290}, 2020.

\bibitem{shlezinger2018hardware}
N.~Shlezinger, Y.~C. Eldar, and M.~R. Rodrigues, ``Hardware-limited task-based
  quantization,'' \emph{{IEEE} Trans. Signal Process.}, vol.~67, no.~20, pp.
  5223--5238, 2019.

\bibitem{shlezinger2018asymptotic}
------, ``Asymptotic task-based quantization with application to massive
  {MIMO},'' \emph{{IEEE} Trans. Signal Process.}, vol.~67, no.~15, pp.
  3995--4012, 2019.

\bibitem{Salamtian19task}
S.~Salamtian, N.~Shlezinger, Y.~C. Eldar, and M.~M{\'e}dard, ``Task-based
  quantization for recovering quadratic functions using principal inertia
  components,'' in \emph{Proc. IEEE Int. Symp. Inf. Theory}, 2019.

\bibitem{yang2020federated}
K.~Yang, T.~Jiang, Y.~Shi, and Z.~Ding, ``Federated learning via over-the-air
  computation,'' \emph{{IEEE} Trans. Wireless Commun.}, vol.~19, no.~3, pp.
  2022--2035, 2020.

\bibitem{sery2020over}
T.~Sery, N.~Shlezinger, K.~Cohen, and Y.~C. Eldar, ``Over-the-air federated
  learning from heterogeneous data,'' \emph{arXiv preprint arXiv:2009.12787},
  2020.

\bibitem{letaief2019roadmap}
K.~B. Letaief, W.~Chen, Y.~Shi, J.~Zhang, and Y.-J.~A. Zhang, ``The roadmap to
  {6G}: {AI} empowered wireless networks,'' \emph{{IEEE} Commun. Mag.},
  vol.~57, no.~8, pp. 84--90, 2019.

\bibitem{kang2019incentive}
J.~Kang, Z.~Xiong, D.~Niyato, S.~Xie, and J.~Zhang, ``Incentive mechanism for
  reliable federated learning: A joint optimization approach to combining
  reputation and contract theory,'' \emph{{IEEE} Internet Things J.}, vol.~6,
  no.~6, pp. 10\,700--10\,714, 2019.

\bibitem{shlezinger2020communication}
N.~Shlezinger, S.~Rini, and Y.~C. Eldar, ``The communication-aware clustered
  federated learning problem,'' in \emph{Proc. IEEE Int. Symp. Inf. Theory},
  2020, pp. 2610--2615.

\bibitem{kang2020reliable}
J.~Kang, Z.~Xiong, D.~Niyato, Y.~Zou, Y.~Zhang, and M.~Guizani, ``Reliable
  federated learning for mobile networks,'' \emph{{IEEE} Wireless Commun.},
  vol.~27, no.~2, pp. 72--80, 2020.

\bibitem{chen2019artificial}
M.~Chen, U.~Challita, W.~Saad, C.~Yin, and M.~Debbah, ``Artificial neural
  networks-based machine learning for wireless networks: A tutorial,''
  \emph{{IEEE} Commun. Surveys Tuts.}, vol.~21, no.~4, pp. 3039--3071, 2019.

\bibitem{cover2012elements}
T.~M. Cover and J.~A. Thomas, \emph{Elements of Information Theory}.\hskip 1em
  plus 0.5em minus 0.4em\relax John Wiley \& Sons, 2012.

\bibitem{wyner1976rate}
A.~Wyner and J.~Ziv, ``The rate-distortion function for source coding with side
  information at the decoder,'' \emph{{IEEE} Trans. Inf. Theory}, vol.~22,
  no.~1, pp. 1--10, 1976.

\bibitem{agrell1998optimization}
E.~Agrell and T.~Eriksson, ``Optimization of lattices for quantization,''
  \emph{{IEEE} Trans. Inf. Theory}, vol.~44, no.~5, pp. 1814--1828, 1998.

\bibitem{ferentios1982tcebycheff}
K.~Ferentios, ``On {Tcebycheff's} type inequalities,'' \emph{Trabajos de
  Estadistica y de Investigacion Operativa}, vol.~33, no.~1, p. 125, 1982.

\bibitem{stich2018local}
S.~U. Stich, ``Local {SGD} converges fast and communicates little,''
  \emph{arXiv preprint arXiv:1805.09767}, 2018.

\bibitem{conway1982voronoi}
J.~Conway and N.~Sloane, ``Voronoi regions of lattices, second moments of
  polytopes, and quantization,'' \emph{{IEEE} Trans. Inf. Theory}, vol.~28,
  no.~2, pp. 211--226, 1982.

\bibitem{zhang2013communication}
Y.~Zhang, J.~C. Duchi, and M.~J. Wainwright, ``Communication-efficient
  algorithms for statistical optimization,'' \emph{The Journal of Machine
  Learning Research}, vol.~14, no.~1, pp. 3321--3363, 2013.

\bibitem{koloskova2020unified}
A.~Koloskova, N.~Loizou, S.~Boreiri, M.~Jaggi, and S.~U. Stich, ``A unified
  theory of decentralized {SGD} with changing topology and local updates,''
  \emph{arXiv preprint arXiv:2003.10422}, 2020.

\bibitem{matlab2020}
\BIBentryALTinterwordspacing
{MathWorks Deep Learning Toolbox Team}, ``Deep learning tutorial series,''
  \emph{MATLAB Central File Exchange}, 2020. [Online]. Available:
  \url{https://www.mathworks.com/matlabcentral/fileexchange/62990-deep-learning-tutorial-series}
\BIBentrySTDinterwordspacing

\bibitem{Guozhong1995NoiseBackprop}
G.~An, ``The effects of adding noise during backpropagation training on a
  generalization performance,'' \emph{Neural computation}, vol.~8, no.~3, pp.
  643--674, 1996.

\end{thebibliography}
	
\end{document}